\documentclass[runningheads]{llncs}

\usepackage{eccv}

\usepackage{eccvabbrv}

\usepackage{graphicx}
\usepackage{booktabs}
\usepackage{multirow}
\usepackage[accsupp]{axessibility}  
\usepackage{wrapfig}

\usepackage{algorithm}
\usepackage{algorithmicx}
\usepackage{algpseudocode}
\usepackage{wrapfig}   

\usepackage{xcolor}

\usepackage{bm}
\usepackage{amsmath}
\DeclareMathOperator*{\argmax}{arg\,max}

\usepackage{siunitx}

\usepackage{hyperref}

\usepackage{orcidlink}

\begin{document}



\title{$T^{3}S$: Think in Thermal Time for Generalizable Crop Mapping from Satellite Image Time Series}

\titlerunning{Generalizable Crop Mapping from Satellite Image Time Series}

\author{Mehmet Ozgur Turkoglu\inst{1}\orcidlink{0000-0003-1446-2778} \and
Sélène Ledain\inst{1}\orcidlink{0009-0002-9321-1933} \and
Jeffrey Zweidler\inst{2}\and \\
Thomas Lauber\inst{1}\orcidlink{0000-0002-3118-432X}  \and
Helge Aasen\inst{1}\orcidlink{0000-0003-4343-0476}}

\authorrunning{Turkoglu et al.}

\institute{Earth Observation of Agroecosystems Team, Agroscope, Switzerland \and
ETH Zurich, Switzerland\\
\email{moturkoglu@gmail.com, selene.ledain@agroscope.admin.ch, zjeffrey@student.ethz.ch, thomas.lauber@agroscope.admin.ch, helge.aasen@agroscope.admin.ch}\\
}

\maketitle

\begin{abstract}
Crop type classification from optical satellite time series remains limited in its ability to generalize across growing seasons, particularly when crop phenology shifts due to inter-annual weather variability. This hampers deployment in operational settings where current-year labels are unavailable. In addition, uncertainty quantification is often overlooked, reducing the reliability of such approaches for practical crop monitoring. Inspired by ecophysiological principles, we introduce Thermal Time-based Temporal Sampling ($T^3S$), a simple, model-agnostic method that replaces calendar time with thermal time. By re-indexing satellite observations by cumulative growing degree days, $T^3S$ aligns phenologically equivalent growth stages across years, reducing temporal redundancy while concentrating on the most biologically informative periods. We evaluate $T^3S$ across three architecturally distinct backbones on (i) SwissCrop, a new country-scale, multi-year Sentinel-2 dataset with paired temperature data that we publicly release, and (ii) the cross-region TimeMatch benchmark spanning Denmark and France. Across these settings, $T^3S$ consistently improves cross-year and cross-region crop classification over several state-of-the-art baselines, including thermal positional encoding, with particularly strong gains in uncertainty calibration, robustness under label scarcity, and early-season prediction, while requiring no architectural modification.
\end{abstract}

\section{Introduction}\label{introduction}

Monitoring agricultural land use is crucial for ensuring the sustainability of global food systems \cite{Gomez2016}. As the world’s population continues to grow, climates become more extreme, and diets evolve, the demand for reliable, affordable and scalable monitoring tools is increasing. Traditional methods, such as field surveys and self-reported data, can be labor-intensive, expensive, and susceptible to errors. In contrast, remote sensing-based approaches offer a faster, more cost-effective, and less subjective alternative for tracking land-use dynamics \cite{weiss2020remote}.

Recent advances in remote sensing and machine learning algorithms have significantly improved the monitoring of agricultural land use. High-resolution, multi-spectral imagery from platforms such as Sentinel-2 provides frequent, detailed observations of croplands at minimal cost. By applying advanced machine learning methods, ranging from conventional ensemble learners like XGBoost \cite{chen2016xgboost} to deep neural networks such as Transformers \cite{vaswani2017attention}, to these time series images, it is possible to capture crop-specific growth dynamics and distinguish among diverse management practices with high precision. These integrated approaches enable automated, scalable mapping of crop types and growth stages, supporting timely, data-driven decisions for sustainable agriculture \cite{turkoglu2023deep}.


%

Modern deep learning methods~\cite{garnot2021panoptic, turkoglu2021crop, metzger2021crop, tarasiou2023vits, russwurm2018multi} excel in classifying crop types from optical satellite image time series. However, these models are typically trained on curated datasets with limited size and/or validated on data collected within a single growing season~\cite{russwurm2017temporal, breizhcrops, schneider2023eurocrops, garnot2021panoptic, turkoglu2021crop, reuss2025eurocropsml}, neglecting inter-annual variability in crop growth driven by climate fluctuations. This limits their applicability across years and regions. Furthermore, their dependence on current-year labeled data for training prevents real-time deployment during the ongoing season. As highlighted by \cite{Roscher2024Better}, developing models with datasets that reflect real-world conditions is essential to further improve the accuracy, generalization, and practical impact of these tools for end users.
Lastly, although these approaches may achieve state-of-the-art accuracy, they rarely incorporate uncertainty quantification, leaving predictive confidence unassessed and often uncalibrated. Ensuring the models are well-calibrated makes them reliable components of agricultural monitoring systems, where miscalibration could otherwise lead to costly resource misallocations or poor risk mismanagement.

\begin{wrapfigure}{r}{0.48\textwidth}
\vspace{-10pt}
    \centering
    \includegraphics[width=.46\textwidth]{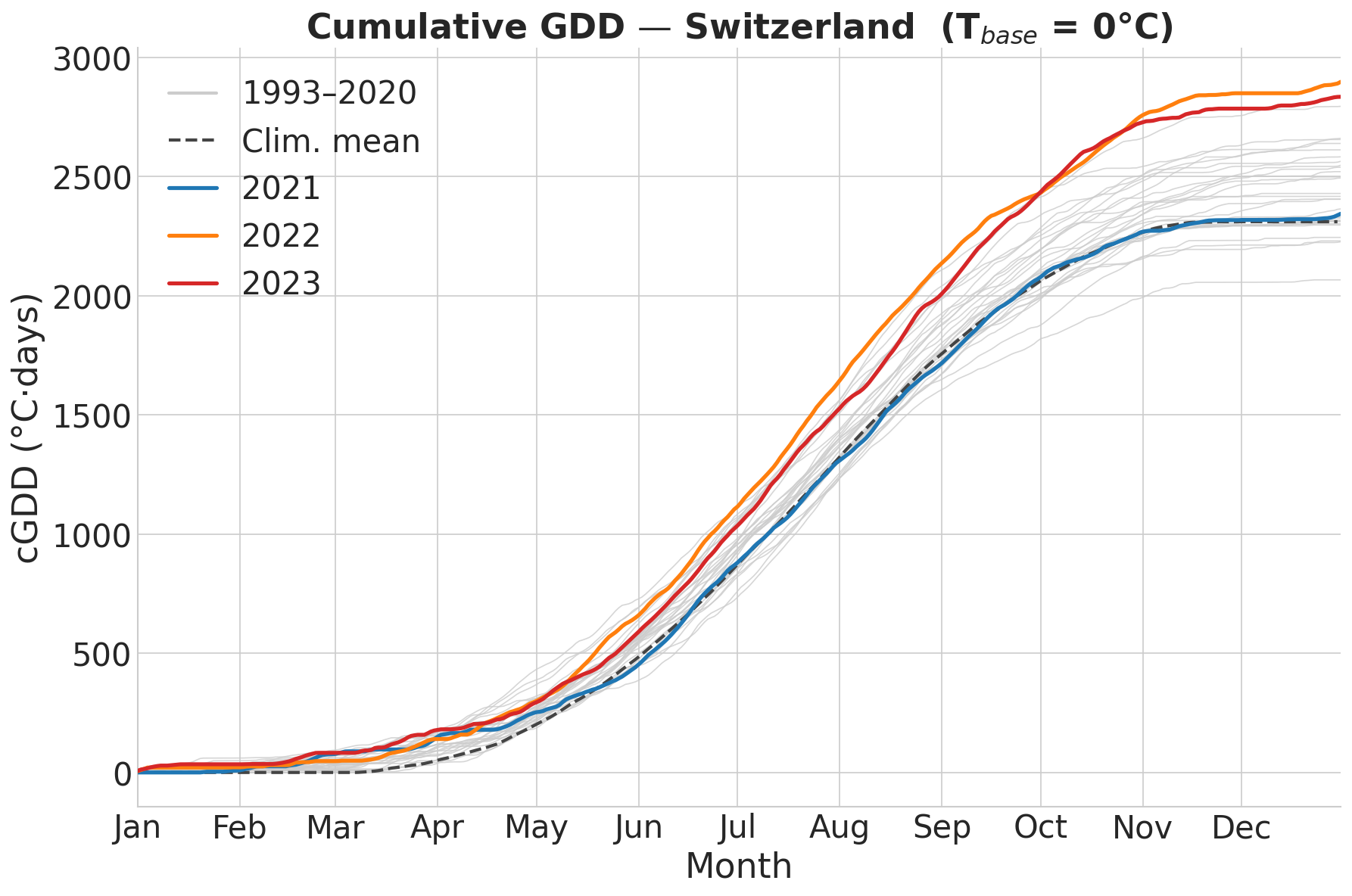}
\caption{Cumulative growing degree days (cGDD) across Switzerland over the last 30 years.}
\label{fig/thermal_calendar}
\vspace{-10pt}
\end{wrapfigure}

\begin{figure}[t]
    \centering
        \includegraphics[width=0.5\textwidth]{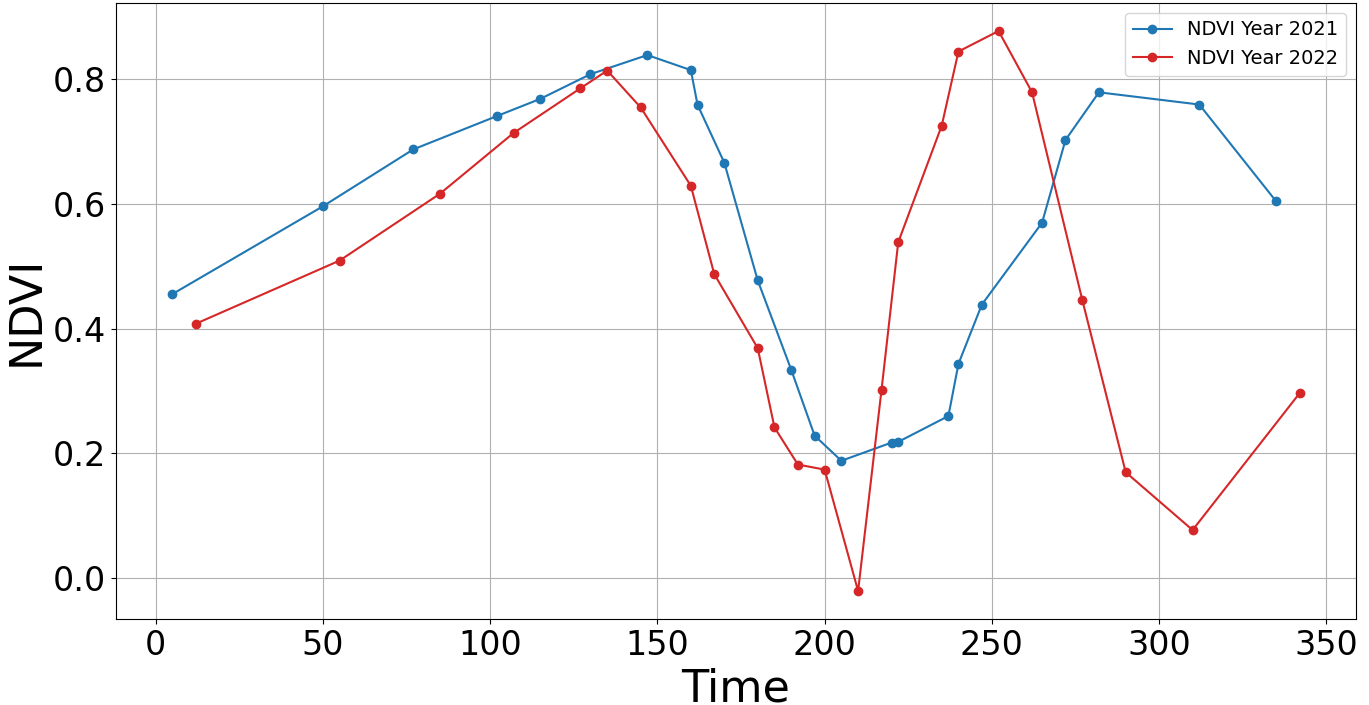}%
        \label{fig:ndvi1}
        \includegraphics[width=0.5\textwidth]{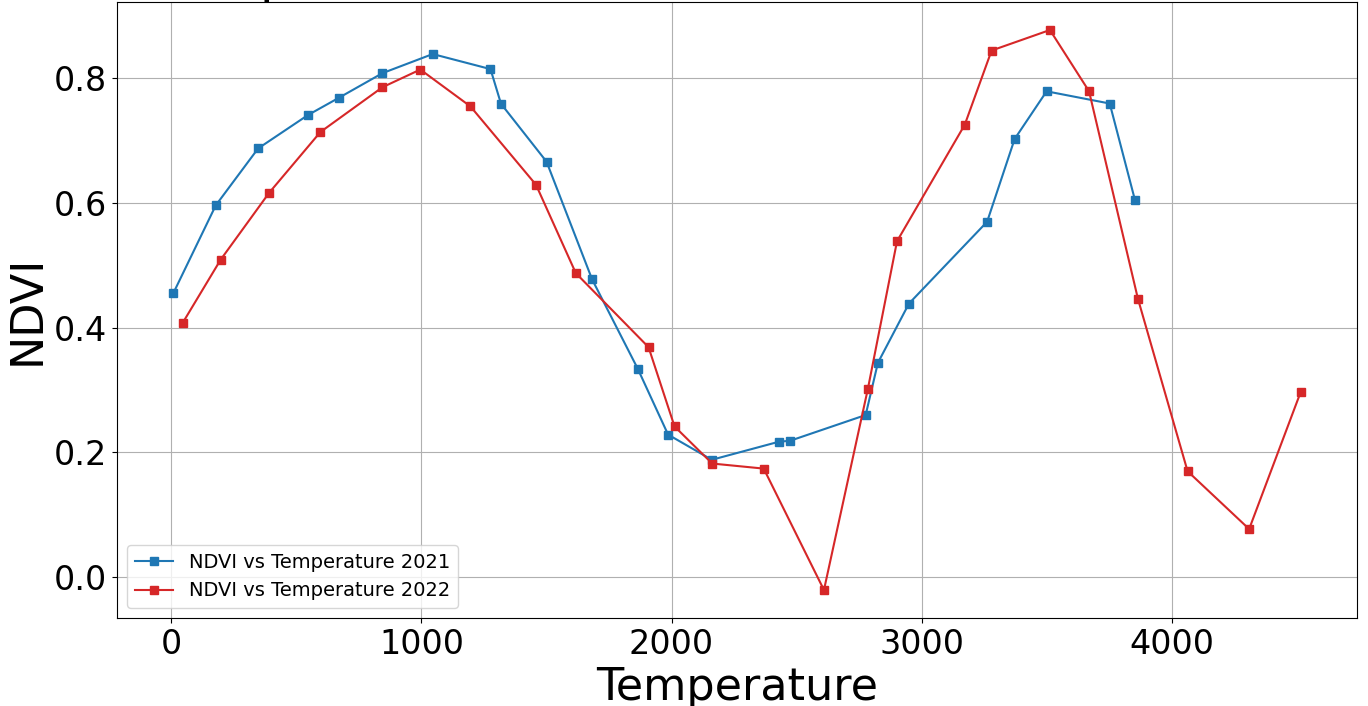}%
        \label{fig:ndvi2}
    \caption{NDVI time series of two parcels with the same crop rotation over two years (2021 and 2022), aligned by calendar date (left) and by cumulative growing degree days (cGDD) (right).}

   \label{fig:ndvi_combined}
\end{figure}

\begin{wrapfigure}{r}{0.6\textwidth}
\vspace{-10pt}
    \centering
    \centering
    \includegraphics[width=0.6
    \textwidth]{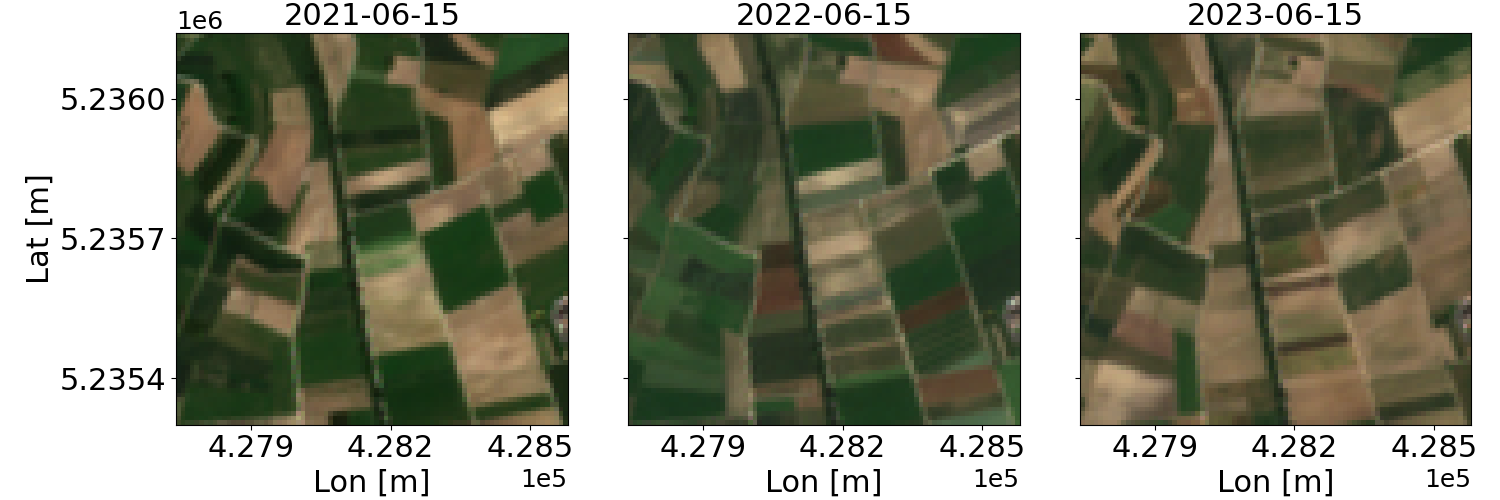}
    \caption{Year-to-year variation of the same agricultural landscape on June 15th from 2021 to 2023.}
    \label{fig:fields_example}
    \vspace{-10pt}
\end{wrapfigure}

Crop development is largely driven by thermal conditions, with temperature playing a
pivotal role in determining growth and phenological development
\cite{mcmaster1997gdd,trudgill2005thermal}. However, temperature can vary substantially from year to year, as illustrated in Figure \ref{fig/thermal_calendar}, which shows cumulative growing degree days (cGDD) in Switzerland across the last 30 years. Consequently, the same crop can reach key growth stages weeks earlier or later depending on the season. This phenological variability is directly reflected in satellite observations, as shown in Figure~\ref{fig:ndvi_combined} (left), which depicts Normalised Difference Vegetation Index (NDVI) time series for a single field over two years, revealing a noticeable shift in the growth trajectory despite the crop type remaining constant.
Figure~\ref{fig:fields_example} further illustrates how the same agricultural landscape can appear markedly different in satellite imagery across different years.

From a modelling perspective, this implies that a model trained on calendar-indexed time series encounters a shifting input distribution across years, as phenologically equivalent growth stages appear at different positions in the sequence. Re-indexing observations by thermal time, rather than calendar days, warps the time axis such that equivalent phenological stages are more consistently aligned irrespective of inter-annual temperature variability, substantially reducing the apparent mismatch in growth cycles (Figure~\ref{fig:ndvi_combined} (right)). While other environmental drivers like precipitation undoubtedly also influence crop growth, their incorporation falls beyond the scope of this study and is reserved for future research.

Motivated by this, we build upon the state-of-the-art attention-based deep learning architecture proposed by \cite{garnot2021panoptic}, and introduce a simple, model-agnostic sampling strategy that incorporates daily average temperature to align the satellite image time series and reduce redundancy and noise. Deep learning models for crop classification typically rely on subsampled time series, as crop development progresses gradually and many satellite observations are temporally redundant~\cite{metzger2021crop}. 
Subsampling is also necessary to reduce sequence length for efficient model training and inference, particularly for attention-based architectures where computational cost scales quadratically with sequence length.

%
%
To better guide this sampling, we reparameterize calendar time into thermal time, using cumulative growing degree days (cGDD) as our temporal reference, and perform uniform subsampling in this thermal‐time domain,  ensuring the model concentrates on the most phenologically informative observations.


%
We evaluate our GDD-informed sampling approach on a multi-year, country-wide dataset spanning three Swiss growing seasons by training on one year and testing on the other two. We benchmark against state-of-the-art baselines, evaluate low-data scenarios using only 10\% of the training samples, assess early-season classification performance, and examine uncertainty calibration to support reliable decision-making. We further assess if our proposed method can improve cross-region generalization between distinct climatic regions using the TimeMatch benchmark \cite{nyborg2022timematch}.
We hypothesize that GDD-informed sampling improves generalization across years and regions, while also boosting uncertainty calibration and robustness under data scarcity and early-season conditions.


In summary, our contributions are as follows:

\begin{itemize}
\item We introduce a simple yet effective model-agnostic approach grounded in the ecophysiological principles of plant growth: Thermal Time-based Temporal Sampling ($T^{3}S$), which enables the integration of temperature data into any machine learning model, irrespective of model architecture.
%
%
\item We benchmark $T^{3}S$ across three architecturally distinct backbones, U-TAE
(CNN with temporal attention), PSE+LTAE (parcel-level pixel-set encoder with lightweight
attention), and Galileo (a pretrained EO foundation model), demonstrating substantial and
consistent gains in predictive accuracy and uncertainty quantification across cross-year and
cross-region experiments, including low-data and early-season settings, and validating its
model-agnostic design.
\item We publicly release SwissCrop, along with code for seamless integration into ML workflows. To our knowledge, it is the first crop dataset that is simultaneously country-scale, multi-year, and paired with temperature data
(Table~\ref{tab:datasets}).
\end{itemize}

\section{Related Work}\label{RW}

Deep learning has revolutionized crop-type classification from satellite time series, with architectures progressively improving in their ability to capture phenological dynamics. Early sequence models based on LSTM \cite{russwurm2017temporal} and convolutional-RNN architectures \cite{turkoglu2021crop} established the importance of temporal modelling, while attention-based approaches subsequently improved the modelling of long-range temporal dependencies, including self-attention \cite{russwurm2020self,garnot2020satellite}, the lightweight Temporal Attention Encoder (L-TAE) \cite{ltae}, and U-Net--integrated attention \cite{garnot2021panoptic}. More recently, Vision Transformer adaptations \cite{bai2023vits} and state-space models like Mamba \cite{qin2024sitsmamba} have further enhanced spatio-temporal representations.

Although crop classification models have become increasingly sophisticated, they rarely provide reliable uncertainty estimates, even though such information is essential for trustworthy deployment. Deep ensembles \cite{lakshminarayanan2017simple} remain the standard for uncertainty quantification, alongside more efficient approximations \cite{gal2016dropout,huang2017snapshot,durasov2021masksembles,Turkoglu2022FiLM-Ensemble:Modulation,halbheer2024lora,turkoglu2026making}. In satellite-based crop monitoring, such stochastic inference yields per-pixel and per-field uncertainty that flags low-confidence regions for targeted ground truthing and more reliable operational deployment, via MC Dropout \cite{tian2024attention,khallaghi2024generalization,liu2024cropsight} and ensembles \cite{baumert2024probabilistic}.

%

Beyond architectural design, a complementary line of work injects agronomic domain knowledge. Some methods encode class structure: ConvSTAR \cite{turkoglu2024hierarchical} and prototype-based supervision \cite{vivien_prototypes} exploit crop class relationships. A more directly relevant thread draws on \emph{thermal time}, a long-standing agronomic construct that summarizes the temperature-driven pace of crop development \cite{mcmaster1997gdd,trudgill2005thermal} and has been used to improve the timeliness and cross-regional transfer of remote-sensing crop forecasts \cite{franch2015timeliness}. Thermal-time positional encoding (TPE) \cite{nyborg2022generalized} builds on this idea by injecting cumulative growing degree days (cGDD) into the network as a positional signal, and is the work most closely related to ours. 
$T^3S$ shares this thermal-time foundation but differs in \emph{where} the signal acts: instead of \emph{encoding}, it drives the \emph{sampling} of observations, removing the misalignment at its source. This keeps it model-agnostic at negligible cost and, unlike TPE, applies even to backbones without a modifiable positional encoding.
A complementary line of work instead makes models robust to temporal misalignment rather than removing it. Data augmentation \cite{nyborg2022timematch,yuan2025empirical} addresses the problem indirectly by expanding the training input space, so models learn to tolerate shifts while the misalignment remains. Dynamic time warping \cite{petitjean2012dtw} explicitly aligns sequences, but requires costly pairwise matching. Deformable prototype learning \cite{vincent2025pixelwise} introduces temporal flexibility through learnable class prototypes, but embeds this robustness in a specific classifier rather than a general input-level representation.
%

\section{Method}\label{method}
%
%


Formally, the objective is to predict a crop type map $\bm{y} \in  {\rm I\!R}^{H\times W\times C}$ from a sequence of input images $\bm{X} = \{\bm{x}_1,\bm{x}_2,..,\bm{x}_L\} \in  {\rm I\!R}^{H\times W\times B}$. $H$ and $W$ are the height and width of the input images, respectively, $B$ is the number of input bands, $L$ is the number of time stamps in the input sequence that can differ from sample to sample, and $C$ is the number of crop types. 
Our method builds on the U-Net with Temporal Attention Encoder (U-TAE) \cite{ronneberger2015u,garnot2021panoptic}, a state-of-the-art deep learning model for dense segmentation of satellite image time series. 
%
%
Before introducing $T^{3}S$, we analyze the effect of temporal resolution and positional encoding in U-TAE to motivate our sampling design. 

\subsection{U-Net with Temporal Attention Encoder (U-TAE)}\label{sec:utae}
%
%

The U-TAE model fuses spatial feature extraction with temporal attention to segment land cover in satellite image time series \(\{\mathbf{x}_1, \mathbf{x}_2, \ldots, \mathbf{x}_L\}\). Each frame \(\mathbf{x}_i\) is passed through a shared 2D convolutional encoder, producing per-frame feature maps. These maps are then aggregated by a self-attention block that employs positional encodings and scaled dot-product attention to weight timestamps according to their learned relevance. The resulting multi-temporal embedding is fed into a U-Net–style decoder with skip connections, yielding high-resolution, temporally adaptive segmentation masks. U-TAE naturally handles sequences of varying length and is optimized via pixel-wise cross-entropy loss against the ground-truth labels. 
For training details, refer to the Appendix~\ref{training_details}. 


\subsection{Preliminary Analysis with U-TAE on PASTIS}
\label{sec:prelim}

Before introducing $T^3S$, we conduct a preliminary analysis on PASTIS~\cite{garnot2021panoptic}
(Appendix~\ref{app:prelim}), an established benchmark introduced alongside U-TAE and independent
of our evaluation data. Since satellite observations are temporally redundant and long sequences
inflate compute, subsampling is both natural and necessary. The analysis shows that (i) uniformly
subsampling to $T=24$ costs only $0.2\%$ absolute accuracy while roughly halving compute
(Table~\ref{tab:pastis_temp}), which is why we fix $T=24$ throughout; and (ii) once observations
are ordered by subsampling, U-TAE relies on relative ordering rather than absolute time, so
modifying the positional encoding adds little (Figure~\ref{fig:pastis}). These findings motivate
selecting a compact set of phenologically informative observations rather than altering the
encoding -- which we realize through $T^3S$.

\begin{figure}[t]
    \centering
    \includegraphics[width=0.9\textwidth]{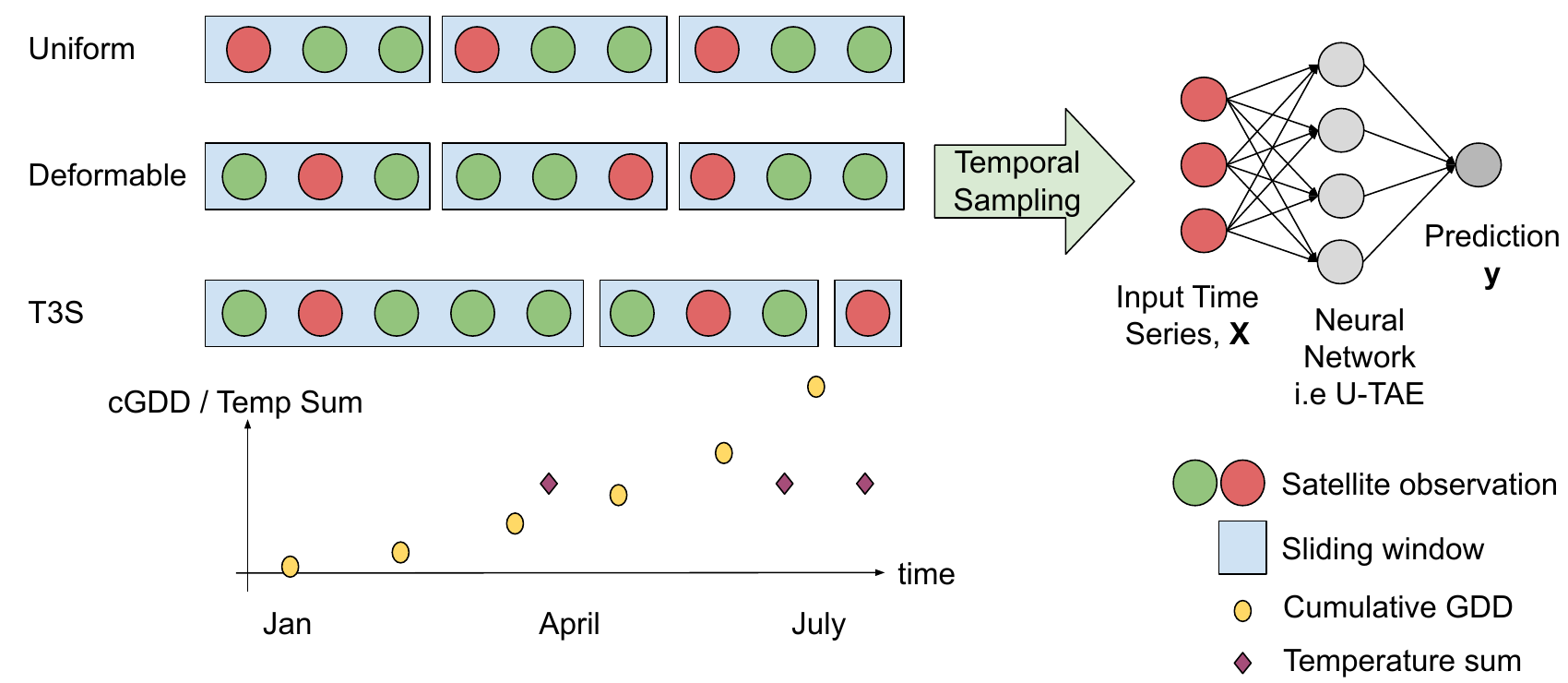}
   \caption{Method overview. The first row shows conventional uniform sampling; the second row shows uniform sampling where the least cloudy observation is chosen in each interval (referred to as “Deformable” baseline); the third row shows the proposed Thermal Time‐based Temporal Sampling ($T^{3}S$). In $T^{3}S$, each square box represents intervals of constant accumulated GDD.}
    \label{fig/method}
\end{figure}

\subsection{$T^{3}S$: Thermal Time-based Temporal Sampling}\label{method:t3s}

\begin{algorithm}[t]
\raggedright
\caption{$T^{3}S$}
\label{alg1}
\begin{algorithmic}[1]
\Function{$T^{3}S$}{$\bm{X}$, cGDD, cloudMask, $T$}
    \State $t_{\min} \gets \min(\text{cGDD})$ \ \ \ \ \# $\bm{X}$'s shape is [L,H,W,B]
    \State $t_{\max} \gets \max(\text{cGDD})$
    \State $\Delta \gets (\,t_{\max} - t_{\min}\,)/T$ \ \ \#$T$ is target temp. length
    \State $selected \gets [\ ]$
    \State $current \gets t_{\min}$
    \While{$|selected| < T$ \textbf{and} $current < t_{\max}$}
        \State $next \gets current + \Delta$
        \State $window \gets \{\,i \mid current \le \text{cGDD}[i] < next\}$
        \If{$window \neq \emptyset$}
            \State $best \gets \displaystyle\argmax_{i \in window}\sum (\text{cloudMask}[i] = 0)$
            \State append $best$ to $selected$
        \EndIf
        \State $current \gets next$
    \EndWhile
    \State sort($selected$)
    \State \Return $\bm{X}$[selected,\,\dots]\; \# the  shape is [T,H,W,B]
\EndFunction
\end{algorithmic}
\end{algorithm}

%
Satellite-based crop monitoring commonly relies on fixed-interval calendar‑day sampling, implicitly assuming that calendar time reflects crop development. 
%
Since phenology is governed by accumulated heat, we introduce Thermal-Time-based Temporal Sampling ($T^{3}S$), a model‑agnostic method that reparametrizes calendar time into thermal time using cumulative growing degree days (cGDD), and performs uniform subsampling in this domain. 


For a given day $i$, GDD is computed as
\begin{equation}
\mathrm{GDD}_i = \max\left(0, \frac{T_{\text{max}, i} + T_{\text{min}, i}}{2} - T_{\text{base}}\right),
\end{equation}
where $T_{\text{max}, i}$ and $T_{\text{min}, i}$ are the daily maximum and minimum temperatures for day $i$, respectively, and $T_{\text{base}}$ is a base threshold below which crop development is considered negligible. 
In our experiments, we use $T_{\text{base}} = 0^\circ$C, a common value for temperate-climate
crops~\cite{mcmaster1997gdd}. Since the crop type is unknown at sampling time, a single crop-agnostic threshold is required; \cite{nyborg2022generalized} adopts the same $0^\circ$C value for this reason.
%
%
We then compute the cumulative GDD (cGDD) over time for day d:
\begin{equation}
\mathrm{cGDD}_d = \sum_{i=1}^{d} \mathrm{GDD}_i.
\end{equation}
To obtain $T$ observations, we partition the cGDD range $\left[\mathrm{cGDD}_{\min}, \mathrm{cGDD}_{\max}\right]$ into $T$ equal intervals and select, within each interval, the least cloudy satellite observation using the Sentinel-2 cloud mask. Algorithm~\ref{alg1} details this procedure, and Figure~\ref{fig/method} illustrates the overall workflow.

In summary, $T^{3}S$ samples observations at comparable thermal stages, reducing temporal redundancy and inter-annual phenological mismatch while remaining fully model-agnostic.


\section{SwissCrop Dataset}\label{sec:dataset}
We introduce SwissCrop, a three-year (2021--2023) country-scale crop dataset covering the entirety of Switzerland, designed to enable rigorous evaluation of cross-year generalization under realistic inter-annual variability.
SwissCrop combines Sentinel-2 Level-2A bottom-of-atmosphere multi-spectral imagery (averaging 110 timestamps per season) with annual crop‐type labels provided at the field level (see Figure~\ref{fig/dataset_overview} in the Appendix).
To enable phenology-aware sampling, we complement the imagery with daily temperature data from MeteoSwiss \cite{federal_office_of_meteorology_and_climatology_meteoswiss_climate_2024}. Temperature data are originally provided at 1~km resolution and resampled to the 10~m Sentinel-2 grid using nearest-neighbor interpolation. These data are used to compute growing degree days (GDD), which serve as the basis for the thermal-time sampling (Section~\ref{method:t3s}).
The dataset comprises 50 distinct crop types, each corresponding to the primary crop grown in a field during a given season.
Although three seasons may appear limited, they span an unusually wide climatic range (Figure~\ref{fig/thermal_calendar}): 2021 was cool and wet, whereas 2022 and 2023 were the two warmest years on record at the time, with annual mean temperatures $+1.6^{\circ}$C and $+1.4^{\circ}$C above the 1991--2020 norm~\cite{meteoswiss_klim2022,meteoswiss_klim2023}. The 2021$\rightarrow$2022 transition alone induced phenological shifts beyond typical inter-annual variability, making SwissCrop a demanding testbed for cross-year generalization despite its three-year span. Reliable country-scale labels became available from 2021 onward.
In contrast to curated benchmarks such as PASTIS \cite{garnot2021panoptic}, SwissCrop reflects real-world agricultural imbalance, following a pronounced long-tail distribution in which a small number of major crops dominate while many others occur only sporadically (see Figure~\ref{fig:label_dist} in the Appendix for the class distribution of SwissCrop).
To our knowledge, SwissCrop is the first crop dataset that is simultaneously country-scale, multi-year, and paired with temperature data (Table~\ref{tab:datasets}), making it uniquely suited to benchmark phenology-aware methods such as $T^{3}S$. The full dataset, including imagery, crop labels, and temperature time series, is publicly released to support reproducible benchmarking of cross-year and phenology-aware models.
For more details, refer to the Appendix~\ref{app:data}.

\section{Experiment}\label{experiment}
We evaluate $T^{3}S$ along three axes: (i) cross-year generalization and uncertainty calibration on the SwissCrop dataset using the U-TAE backbone, (ii) robustness under label scarcity and early-season conditions, and (iii) cross-region generalization using the TimeMatch dataset~\cite{nyborg2022timematch} with the PSE+LTAE backbone~\cite{garnot2020satellite}. For cross-year experiments, we use a leave-one-year-out scheme over the 2021--2023 SwissCrop seasons: for each of the six folds, models are trained on one season and evaluated on the remaining two. To assess robustness, we additionally perform (i) a low-data experiment using only 10\% of the labels, and (ii) early-season evaluation by truncating test sequences at the 50\textsuperscript{th} and 75\textsuperscript{th} percentiles (end of June and end of September, respectively). We measure classification performance using overall accuracy, intersection-over-union (IoU, i.e., overall IoU), and mean IoU (mIoU), which is the macro-averaged per-class IoUs. Uncertainty calibration is quantified using expected calibration error (ECE), which measures the discrepancy between predicted confidence and actual accuracy, negative log-likelihood (NLL), and Brier score.
Code and pretrained checkpoints will be released.

\begin{table*}[t]
\centering

\caption{Cross-year performance on SwissCrop across six evaluation settings
(train/test year in the first two columns). Best per setting and metric in \textbf{bold}, second best \underline{underlined}.}

\label{tab:performance}
\resizebox{1\linewidth}{!}{
\begin{tabular}{c c | c | c c c | c c c}
\toprule
\textbf{Train} & \textbf{Test} & \textbf{Method} & \textbf{Accuracy \% ($\uparrow$)} & \textbf{mIoU \% ($\uparrow$)} & \textbf{IoU \% ($\uparrow$)} & \textbf{ECE \% ($\downarrow$)} & \textbf{NLL ($\downarrow$)} & \textbf{Brier Score ($\downarrow$)} \\
\midrule

\multirow{5}{*}{2021} & \multirow{5}{*}{2022} & U-TAE~\cite{garnot2021panoptic}          & 70.1  & 17.2  & 53.9  & 5.3          & 0.943          & 0.412          \\
                      &                       & +MC-Dropout~\cite{gal2016dropout}        & 71.3  & 17.4  & 55.4  & \underline{3.2} & 0.911          & 0.402          \\
                      &                       & +TPE (Sinusoidal)~\cite{nyborg2022generalized}& 72.5  & 18.6  & 56.9  & 4.5          & 0.854          & 0.385          \\
                      &                       & +Deformable Sampling                     & \underline{73.8} & \underline{20.5} & \underline{58.5} & 4.4          & \underline{0.836} & \underline{0.371} \\
                      &                       & $T^{3}S$ (ours)                                      & \textbf{76.0} & \textbf{21.6} & \textbf{61.3} & \textbf{0.4} & \textbf{0.720} & \textbf{0.329} \\

\midrule
\multirow{5}{*}{2021} & \multirow{5}{*}{2023} & U-TAE~\cite{garnot2021panoptic}          & 71.4  & 16.7  & 55.6  & 5.2          & 0.947          & 0.399          \\
                      &                       & +MC-Dropout~\cite{gal2016dropout}        & 72.5  & 17.0  & 56.8  & 3.7          & 0.918          & 0.394          \\
                      &                       & +TPE (Sinusoidal)~\cite{nyborg2022generalized}& 71.3  & 15.5  & 55.4  & 3.5          & 0.925          & 0.401          \\
                      &                       & +Deformable Sampling                     & \underline{74.6} & \underline{18.9} & \underline{59.6} & \underline{3.2} & \underline{0.822} & \underline{0.359} \\
                      &                       & $T^{3}S$ (ours)                                      & \textbf{77.6} & \textbf{21.0} & \textbf{63.5} & \textbf{0.5} & \textbf{0.691} & \textbf{0.307} \\

\midrule
\multirow{5}{*}{2022} & \multirow{5}{*}{2021} & U-TAE~\cite{garnot2021panoptic}          & 70.8  & 16.2  & 54.8  & 3.5          & 0.936          & 0.408          \\
                      &                       & +MC-Dropout~\cite{gal2016dropout}        & 71.2  & 16.2  & 55.3  & \underline{2.4} & 0.913          & 0.399          \\
                      &                       & +TPE (Sinusoidal)~\cite{nyborg2022generalized}& 72.8  & 18.2  & 57.2  & 7.7          & 1.022          & 0.403          \\
                      &                       & +Deformable Sampling                     & \underline{76.8} & \textbf{20.4} & \underline{62.3} & 4.3          & \underline{0.833} & \underline{0.339} \\
                      &                       & $T^{3}S$ (ours)                                      & \textbf{77.3} & \underline{20.1} & \textbf{63.0} & \textbf{1.3} & \textbf{0.697} & \textbf{0.310} \\

\midrule
\multirow{5}{*}{2022} & \multirow{5}{*}{2023} & U-TAE~\cite{garnot2021panoptic}          & 72.1  & 17.4  & 56.4  & 4.1          & 0.875          & 0.389          \\
                      &                       & +MC-Dropout~\cite{gal2016dropout}        & 72.8  & 17.2  & 57.2  & \underline{2.8} & \underline{0.853} & \underline{0.381} \\
                      &                       & +TPE (Sinusoidal)~\cite{nyborg2022generalized}& 63.9  & 14.2  & 46.9  & 12.3          & 1.374          & 0.525          \\
                      &                       & +Deformable Sampling                     & \underline{76.0} & \underline{20.8} & \underline{61.4} & 7.3          & 0.936          & \underline{0.363} \\
                      &                       & $T^{3}S$ (ours)                                      & \textbf{76.8} & \textbf{20.9} & \textbf{62.4} & \textbf{1.9} & \textbf{0.708} & \textbf{0.318} \\

\midrule
\multirow{5}{*}{2023} & \multirow{5}{*}{2021} & U-TAE~\cite{garnot2021panoptic}          & 71.6  & 17.6  & 55.8  & 6.6          & 0.913          & 0.398          \\
                      &                       & +MC-Dropout~\cite{gal2016dropout}        & 71.6  & 17.6  & 55.7  & \underline{4.0} & \underline{0.889} & 0.391 \\
                      &                       & +TPE (Sinusoidal)~\cite{nyborg2022generalized}& 71.8  & 16.3  & 56.0  & 9.8          & 0.987          & 0.413          \\
                      &                       & +Deformable Sampling                     & \underline{75.3} & \underline{20.5} & \underline{60.4} & 10.5          & 0.951          & \underline{0.379} \\
                      &                       & $T^{3}S$ (ours)                                      & \textbf{77.1} & \textbf{23.2} & \textbf{62.8} & \textbf{1.9} & \textbf{0.686} & \textbf{0.307} \\

\midrule
\multirow{5}{*}{2023} & \multirow{5}{*}{2022} & U-TAE~\cite{garnot2021panoptic}          & 71.2  & 18.4  & 55.3  & 5.9          & 0.957          & 0.410          \\
                      &                       & +MC-Dropout~\cite{gal2016dropout}        & 72.0  & 18.6  & 56.2  & \underline{4.3} & 0.929 & 0.400 \\
                      &                       & +TPE (Sinusoidal)~\cite{nyborg2022generalized}& 75.3  & 19.6  & 60.4  & 8.5          & 0.929          & 0.366          \\
                      &                       & +Deformable Sampling                     & \underline{75.8} & \underline{21.9} & \underline{61.0} & 8.1          & \underline{0.827} & \underline{0.355} \\
                      &                       & $T^{3}S$ (ours)                                      & \textbf{76.9} & \textbf{22.3} & \textbf{62.5} & \textbf{0.6} & \textbf{0.687} & \textbf{0.311} \\

\midrule
\midrule
\multirow{5}{*}{Average} & \multirow{5}{*}{}  & U-TAE~\cite{garnot2021panoptic}                    &  $71.2\pm0.6$  &  $17.3\pm0.7$  &  $55.3\pm0.8$  &  $5.1\pm1.2$  &  $0.929\pm0.028$  &  $0.403\pm0.008$  \\
                         &                    & +MC-Dropout~\cite{gal2016dropout}           &  $71.9\pm0.6$  &  $17.3\pm0.7$  &  $56.1\pm0.7$  &  $\underline{3.4}\pm0.7$  &  $0.902\pm0.029$  &  $0.395\pm0.007$  \\
                         &                    & +TPE (Sinusoidal)~\cite{nyborg2022generalized}           &  $71.3\pm3.5$  &  $17.1\pm1.9$  &  $55.5\pm4.1$  &  $7.7\pm2.9$  &  $1.015\pm0.156$  &  $0.416\pm0.053$  \\
                         &                    & +Deformable Sampling  &  $\underline{75.4}\pm1.0$  &  $\underline{20.5}\pm0.9$  &  $\underline{60.5}\pm1.2$  &  $6.3\pm2.5$  &  $\underline{0.868}\pm0.045$  &  $\underline{0.361}\pm0.014$  \\
                         &                    & $T^{3}S$ (ours)                 &  $\textbf{77.0}\pm0.5$  &  $\textbf{21.5}\pm1.0$  & $\textbf{62.6}\pm0.7$  &  $\textbf{1.1}\pm0.6$  &  $\textbf{0.698}\pm0.012$  &  $\textbf{0.314}\pm0.008$  \\
\bottomrule
\end{tabular}
}
\end{table*}

\subsection{Baselines}

To isolate the effect of thermal-time alignment, we compare $T^{3}S$ to four relevant baselines to quantify gains in predictive performance and uncertainty calibration. \textbf{U-TAE} (Section \ref{sec:utae}) serves as our primary reference, with $T=24$ uniformly sampled observations per season (approximately one image every fifteen days). 
\textbf{U-TAE+MC‐Dropout} activates dropout layers at test time to approximate a Bayesian posterior, yielding pixel‐level uncertainty estimates without changing the input dates. 
 %
%
\textbf{U-TAE+TPE} incorporates thermal-time positional encodings (TPE) based on cGDD~\cite{nyborg2022generalized}, providing phenological context through the temporal encoding rather than through the sampled observations. We use the sinusoidal TPE variant for SwissCrop and the RNN-based TPE variant for TimeMatch, as these are the strongest TPE variants for the respective experimental settings.
Lastly, \textbf{U-TAE+\nobreak Deformable Sampling} isolates the cloud-aware selection component of $T^{3}S$ by selecting the least-cloudy observation within each calendar interval, improving data quality while preserving the original calendar-time grid. 
%
We further compare against 
\textbf{Galileo}~\cite{tseng2025galileo}, a state-of-the-art pretrained Earth observation foundation model with a fully attention-based architecture, identified as the top-performing foundation model for agricultural tasks~\cite{nedungadi2025general} (nano variant, due to GPU memory constraints).
For cross-region experiments, we adopt the \textbf{PSE+LTAE} backbone~\cite{garnot2020satellite} following the TimeMatch protocol~\cite{nyborg2022timematch}.

\subsection{Cross‐Year Generalization and Uncertainty Estimation}

$T^{3}S$ consistently outperforms all baselines in cross‐year generalization (see Table \ref{tab:performance}), improving overall accuracy by $\sim$6\% over U-TAE, with gains in IoU and mean IoU. Importantly, it also yields substantially better-calibrated predictions, reducing ECE by roughly fourfold, approaching perfect calibration, while lowering both NLL and Brier score, indicating sharper and more reliable uncertainty estimates. Most crop categories benefit from $T^{3}S$ (Figure~\ref{fig:classwise} in Appendix); only the rarest classes (e.g., hop) remain at zero accuracy due to insufficient training samples. Meadow accuracy drops slightly, reflecting U-TAE’s tendency to overpredict dominant classes and incur more false positives. 
%
%
%

By comparison, across the baselines, U-TAE+Deformable Sampling shows the highest predictive accuracy due to cloud filtering but at the expense of uncertainty quality, exhibiting higher ECE. 
While U-TAE+MC‐Dropout only modestly boosts accuracy relative to vanilla U-TAE, it achieves the second‐best calibration (in terms of ECE). However, note that MC-Dropout is specifically designed to improve uncertainty calibration.
U-TAE+TPE injects phenological context through thermal-time positional encodings, but this yields only a marginal accuracy gain while degrading calibration. Appendix~\ref{app:prelim} analyzes why encoding-based signals add little once observations are ordered, and Section~\ref{sec:svse} isolates sampling from encoding in a controlled comparison against $T^{3}S$.
Overall, these results highlight that aligning observations by thermal time via temporal sampling, combined with cloud‐aware selection, yields the most robust and trustworthy cross‐year crop mapping models.

%

\subsection{Comparison to State-of-the-Art}
%

Table~\ref{tab:sota} compares $T^{3}S$ with Galileo~\cite{tseng2025galileo},
a large pretrained Earth observation foundation model, using U-TAE as a
task-specific state-of-the-art architecture for satellite image time-series crop
mapping. $T^{3}S$+U-TAE achieves the best performance on every metric. The
comparison with Galileo highlights the importance of phenological alignment:
despite its scale and pretraining, Galileo does not match the vanilla U-TAE
baseline in accuracy. This suggests that, for cross-year crop mapping, aligning
observations with crop development can be more important than model capacity
alone, consistent with recent findings that geospatial foundation models remain
immature for agricultural downstream tasks and often underperform task-specific
approaches in cross-domain settings~\cite{ma2025harvesting}.

Applying $T^{3}S$ to Galileo further confirms that the method is architecture
agnostic. The same low-cost preprocessing step, requiring only readily available
temperature data, improves both accuracy and calibration for Galileo, closing
most of the gap to $T^{3}S$+U-TAE and achieving what large-scale pretraining
alone does not. Galileo's residual ECE nevertheless remains higher than that of
$T^{3}S$+U-TAE ($5.8\%$ vs.\ $1.1\%$), suggesting that foundation models may
still require dedicated calibration beyond input-level phenological alignment.

\begin{table}[h]
\centering
\caption{Comparison to state-of-the-art on SwissCrop, averaged over six cross-year folds.
Best per metric in \textbf{bold}, second best \underline{underlined}.}
\label{tab:sota}
\begin{tabular}{c | c | c | c}
\toprule
\textbf{Method} & \textbf{Acc(\%) ($\uparrow$)} & \textbf{IoU(\%) ($\uparrow$)}  &  \textbf{ECE(\%) ($\downarrow$)}  \\
\midrule
  U-TAE~\cite{garnot2021panoptic} & $71.2\pm0.6$ & $55.3\pm0.8$ & $\underline{5.1}\pm1.2$    \\

  Galileo~\cite{tseng2025galileo} & $70.1\pm2.7$ & $54.0\pm3.2$ & $8.8\pm2.3$ \\  
  \midrule
   $T^{3}S$ (U-TAE)  & $\textbf{77.0}\pm0.5$ & $\textbf{62.6}\pm0.7$ & $\textbf{1.1}\pm0.6$ \\  
  $T^{3}S$ (Galileo)    & \underline{74.7} $\pm$ 2.3 & \underline{59.7} $\pm$ 2.9 & {5.8} $\pm$ 1.7 \\

\bottomrule
\end{tabular}

\end{table}

\subsection{Qualitative Comparison}

\begin{figure}[h]
    \centering
    \includegraphics[width=1.0\textwidth]{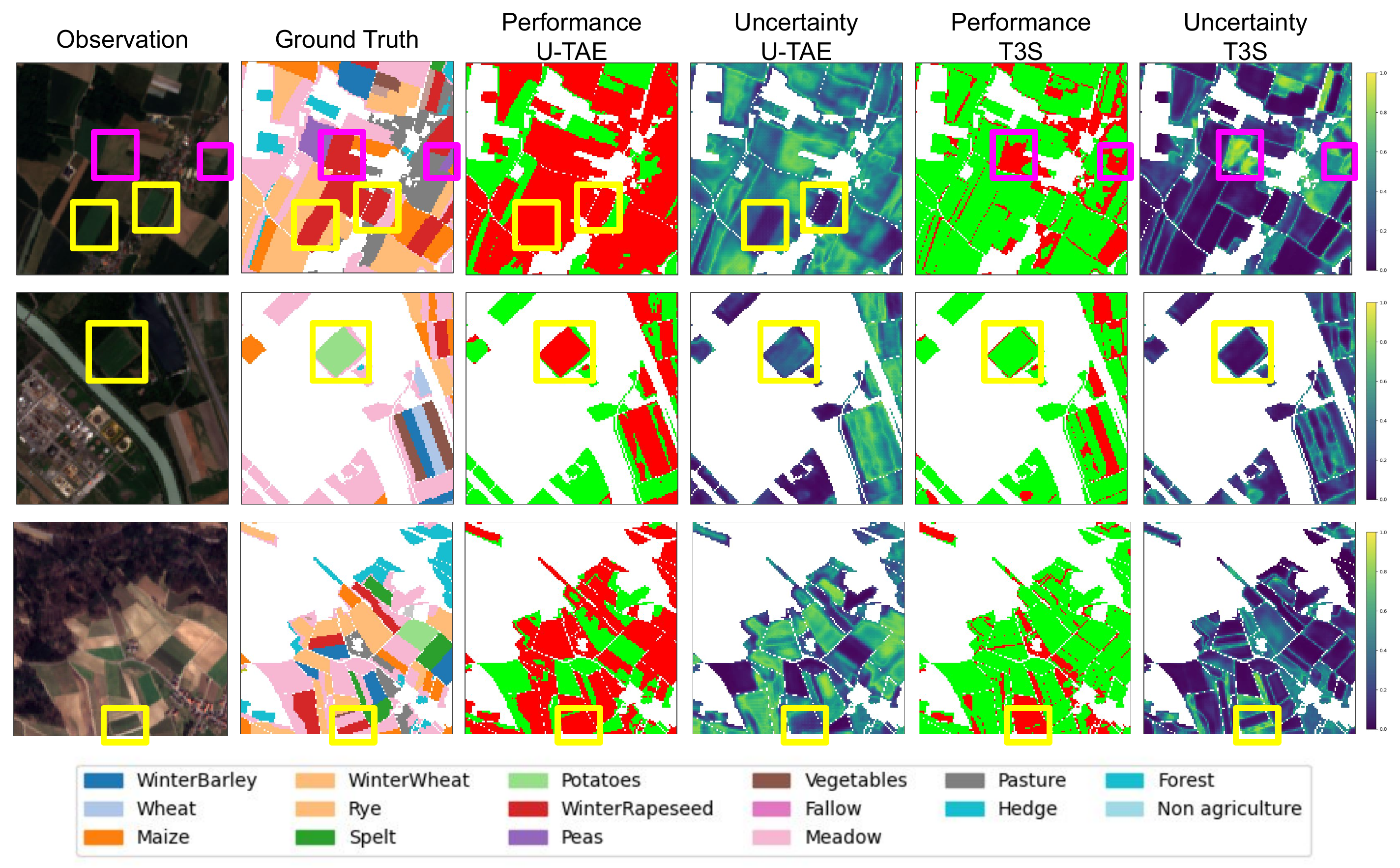}
    \caption{Qualitative comparison between $T^{3}S$ and U-TAE. In the performance panel, green pixels denote correct predictions, red pixels denote errors. In the uncertainty panel, dark blue pixels indicate low uncertainty, while yellow pixels indicate high uncertainty. For best visibility, view the figure zoomed on-screen.}
    \label{fig:qualitative}
\end{figure}

The quantitative improvements are further supported by the qualitative examples in  Figure~\ref{fig:qualitative}, which contrasts $T^{3}S$ and U-TAE on both classification and uncertainty estimation, where pixel-wise uncertainty is defined as $1 - \max(\mathrm{softmax}) \in [0,1]$.
In the first example, both models misclassify the highlighted region, but U-TAE remains overconfident in its incorrect label while $T^{3}S$ appropriately elevates uncertainty around the error. 
%
%
In the second example, $T^{3}S$ confines its errors to the fuzzy edges of the highlighted field. This is understandable given that farmers hand-draw field polygons and 10~m pixels often cover field boundaries. The $T^{3}S$ uncertainty map correctly shows a pronounced uncertainty hotspot along those misclassified borders. In contrast, U-TAE not only mislabels the entire field but also produces a diffuse, misaligned uncertainty map that neither highlights its own errors nor respects the actual field geometry.
In the third example, $T^{3}S$ again flags its error with a pronounced uncertainty hotspot, even revealing a likely ground-truth annotation mistake, revealed by comparing the observation with the ground truth. Again, U-TAE provides no meaningful uncertainty signal for this field. These examples illustrate that $T^{3}S$ not only reduces classification errors but also delivers well-calibrated uncertainty estimates that expose both model and dataset inconsistencies.


\subsection{Robustness under Operational Constraints}

We assess the robustness of $T^{3}S$ under two challenging conditions relevant to operational deployment: early-season classification, where only partial time series are available, and label scarcity, reflecting scenarios in which only limited annotated data are available.

\paragraph{\textbf{Early‐Season Classification.}}

\begin{figure}[h]
    \centering
    \includegraphics[width=0.25\textwidth]{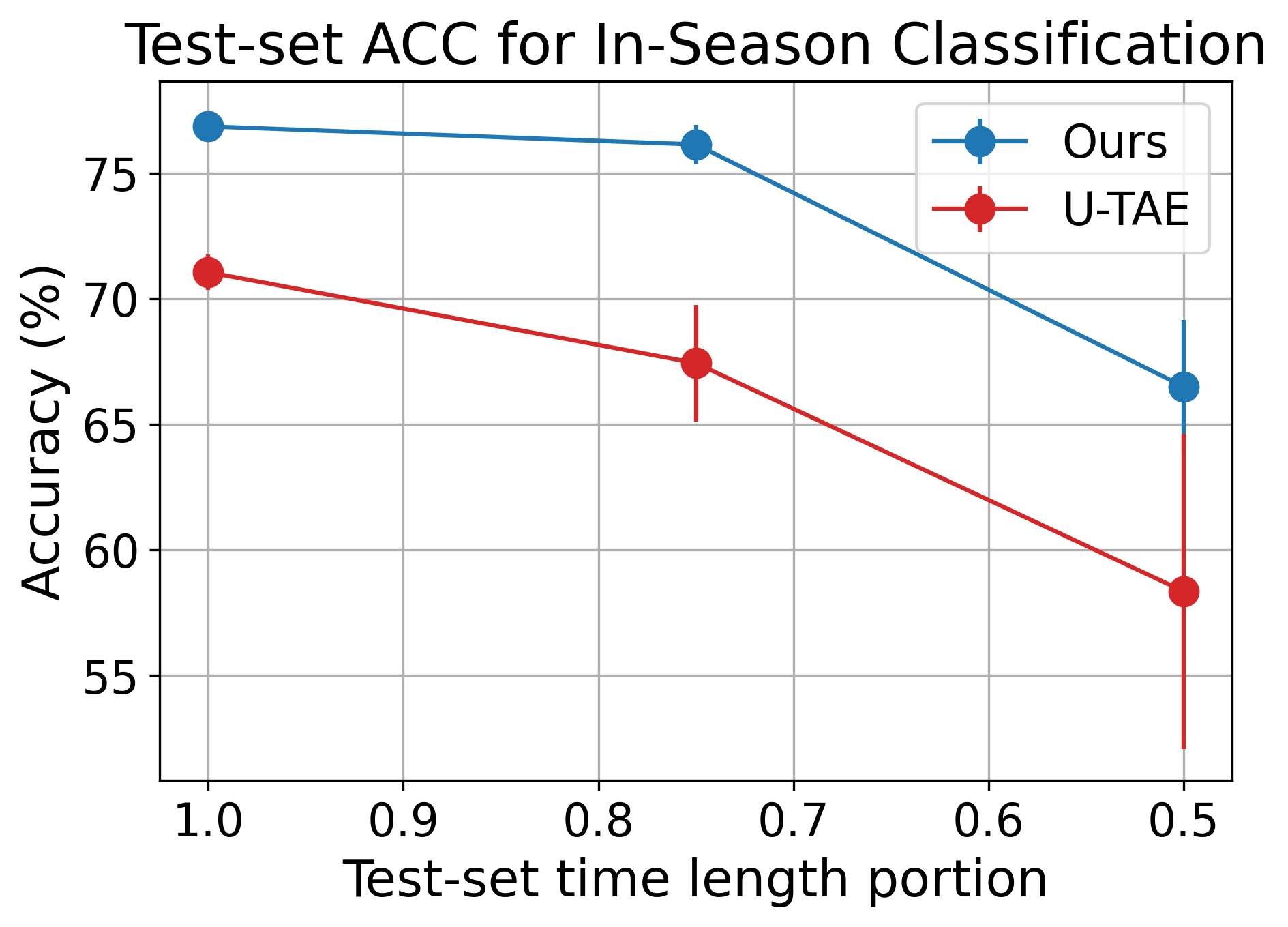}%
    \includegraphics[width=0.25\textwidth]{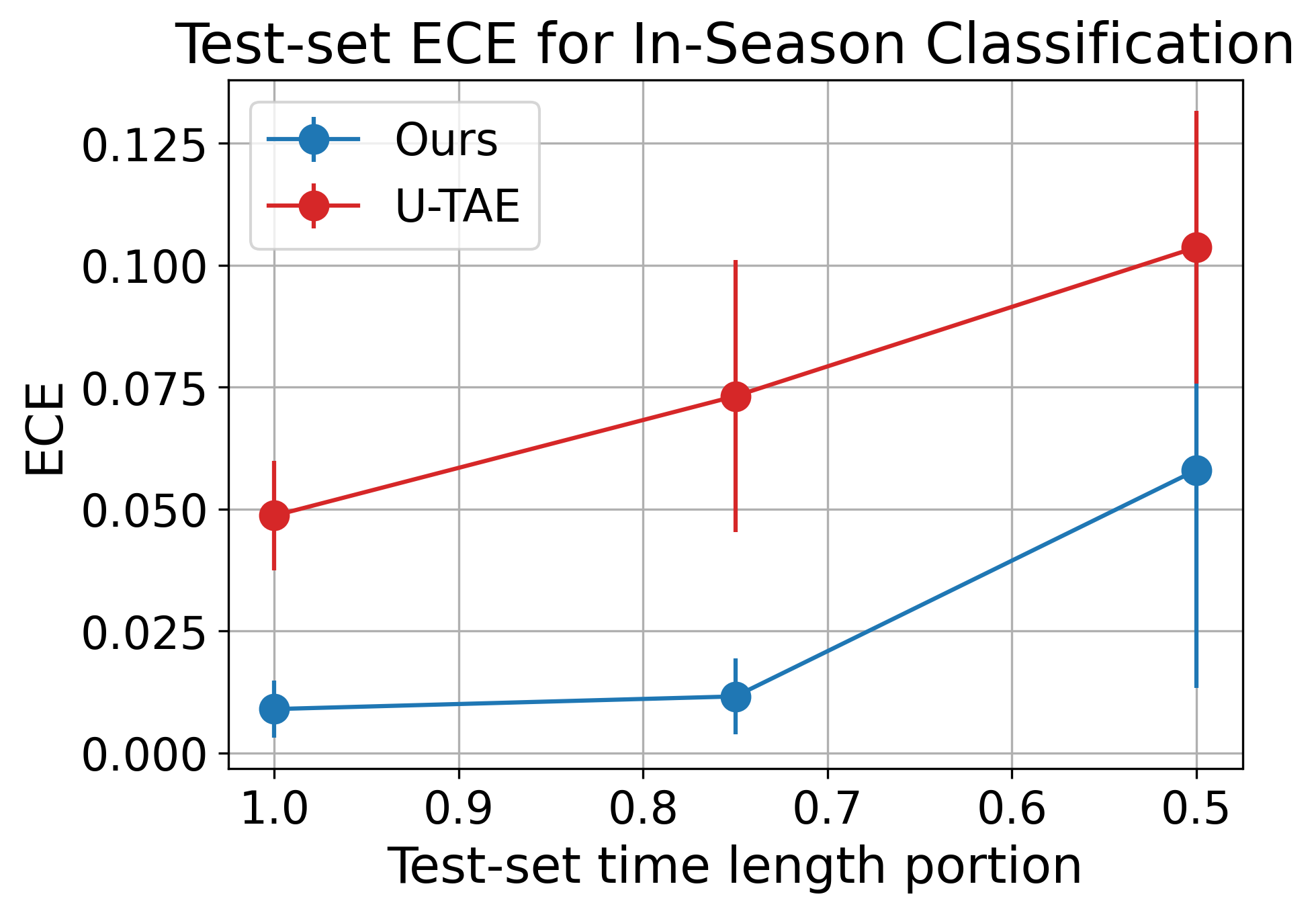}%
    \includegraphics[width=0.25\textwidth]{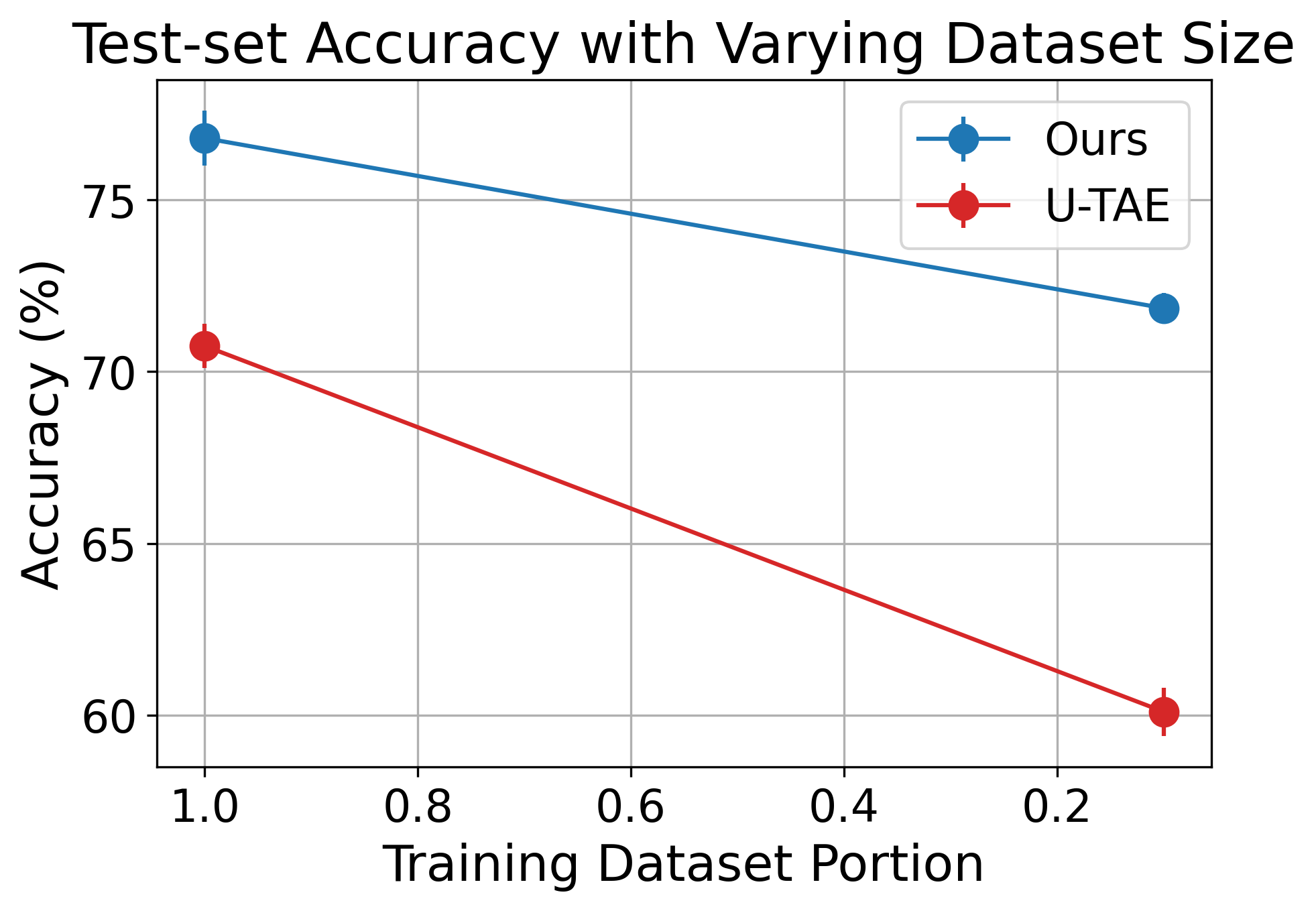}%
\includegraphics[width=0.25\textwidth]{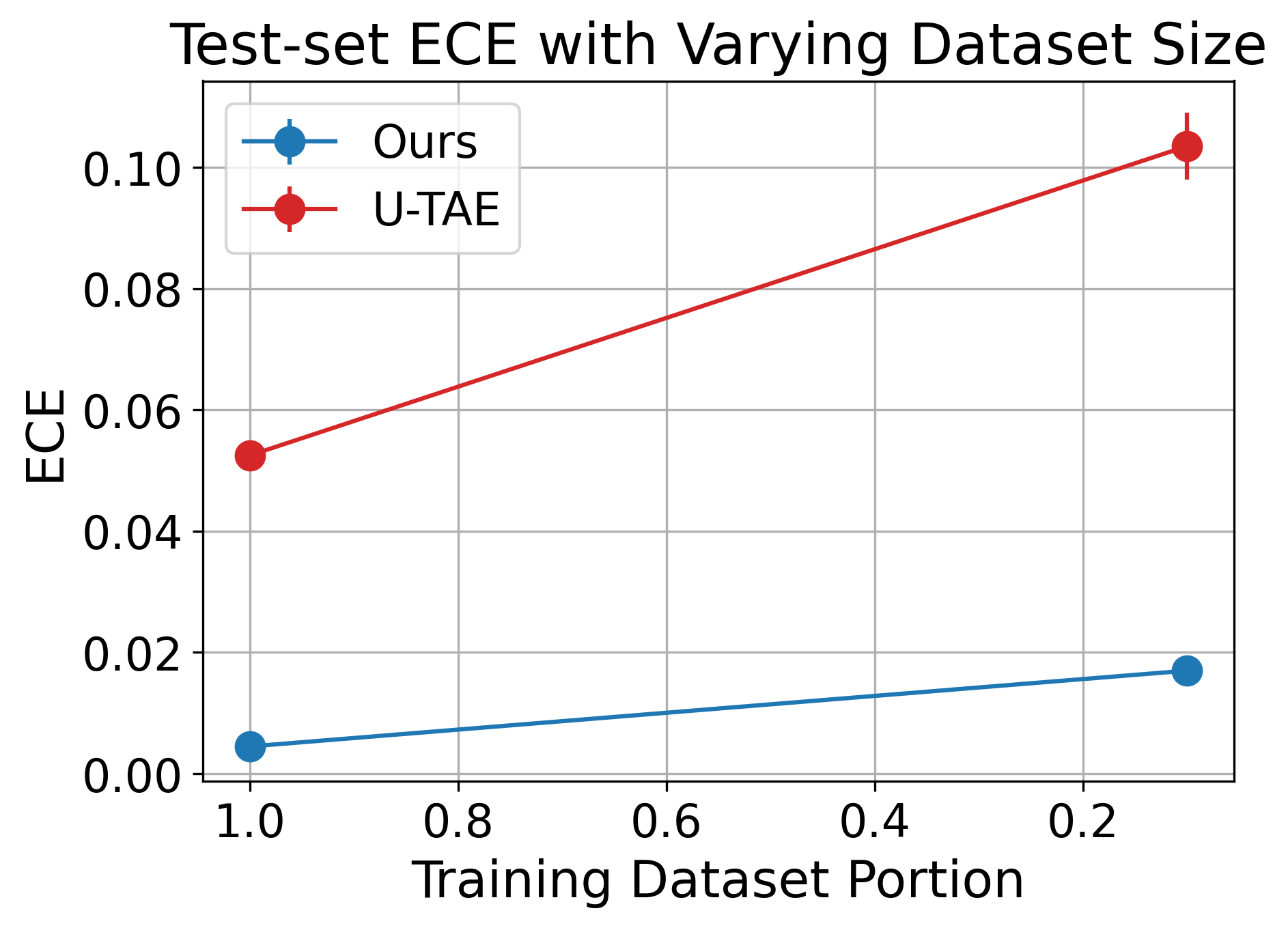}%
\caption{Robustness evaluation averaged over six folds. \textbf{(a,b):} In-season accuracy and ECE at 75th- and 50th-percentile temporal cutoffs (end of September and June). \textbf{(c,d):} Accuracy and ECE when training on 10\% of labeled data.}

    \label{fig:early}
\end{figure}

We truncate the test season at two percentiles, namely 75\textsuperscript{th} (end of September) and 50\textsuperscript{th} (end of June), and perform a six-fold experiment in which both $T^{3}S$ and U-TAE are trained on full-season data and tested on the truncated time series. 
Figure~\ref{fig:early} shows that both models experience a decline in in-season performance, but $T^{3}S$ maintains relatively more stable accuracy, suggesting it can effectively leverage phenological information. At the 75\textsuperscript{th} percentile cutoff, while U-TAE’s performance drops sharply, $T^{3}S$ remains virtually unchanged. Moreover, at the 50\textsuperscript{th} percentile cutoff, $T^{3}S$ still achieves a similar level of accuracy and better uncertainty calibration than U-TAE does at the 75\textsuperscript{th} percentile cutoff. 
These results underscore the practical suitability of $T^{3}S$ for operational in-season crop monitoring, where both timeliness and reliable uncertainty estimates are critical.

\paragraph{\textbf{Low Data Regime.}}
To assess robustness under label scarcity, we conducted a six-fold experiment using only 10\% of the training data, randomly sampled once and held constant across methods, and evaluated both $T^{3}S$ and U-TAE on the full test set. As shown in Figure~\ref{fig:early}, both models incur drops in overall accuracy and increases in ECE when labels are limited, but U-TAE’s performance deteriorates far more rapidly. Remarkably, $T^{3}S$ trained on just 10 \% of the data still outperforms U-TAE trained on the full dataset, and its calibration remains substantially more reliable.

We further analyze the sensitivity of $T^{3}S$ to temporal sequence length in Appendix~\ref{app:temp}, showing that reducing from 24 to 12 timestamps causes only a minor performance drop for $T^{3}S$, while U-TAE degrades more substantially.

\subsection{Cross-Region Generalization with TimeMatch}

Lastly, we evaluate whether $T^{3}S$ improves model transferability across distinct climatic regions. For this, we focus on the two most climatically contrasting subregions in the TimeMatch dataset \cite{nyborg2022timematch}, spanning the South of France and Denmark. We substitute the standard calendar-based sampling with $T^{3}S$ in the PSE+LTAE backbone \cite{garnot2020satellite} and compare our results to the RNN-based Temporal Positional Encoding from \cite{nyborg2022generalized} as the baseline for phenology-aware embeddings.

%
As shown in Table~\ref{tab:tpe}, $T^{3}S$ improves both accuracy and calibration in either transfer setting, significantly outperforming both the standard PSE+LTAE and the RNN-based Thermal Positional Encoding (TPE) approach. These results demonstrate that temperature-informed sampling can also improve generalization across regions, yielding more reliable and phenology-aligned representations than positional encoding alone. In this experiment, results are averaged over five independent runs with different seeds.

\begin{table}[h]
\centering
\caption{Cross-region generalization comparison with TimeMatch. The first three rows show South of France $\rightarrow$ Denmark, while the last three rows show Denmark $\rightarrow$ South of France. Best score for each metric in \textbf{bold}, second best \underline{underlined}.}
\label{tab:tpe}
\begin{tabular}{c | c | c | c }
\toprule
\textbf{Method} & \textbf{Accuracy(\%) ($\uparrow$)} & \textbf{F1(\%) ($\uparrow$)}  &  \textbf{ECE(\%) ($\downarrow$)}  \\
\midrule
  PSE+LTAE~\cite{garnot2020satellite} & $\underline{31.6}\pm3.3$ & $\underline{31.8}\pm3.6$ & $\underline{37.9}\pm3.3$    \\
  + TPE (RNN)~\cite{nyborg2022generalized} & $27.3\pm4.5$ & $26.4\pm5.4$ & $44.0\pm5.2$  \\   
   $T^{3}S$ (PSE+LTAE) & $\textbf{38.6}\pm2.1$ & $\textbf{36.4}\pm2.7$ & $\textbf{31.0}\pm2.8$  \\  

\midrule
  PSE+LTAE~\cite{garnot2020satellite} & $36.5\pm2.2$ & $36.3\pm3.1$ & $37.3\pm2.1$     \\
  + TPE (RNN)~\cite{nyborg2022generalized} & $\underline{47.4}\pm5.9$ & $\underline{49.7}\pm6.6$ & $\underline{26.3}\pm7.4$     \\   
   $T^{3}S$ (PSE+LTAE) & $\textbf{57.7}\pm6.8$ & $\textbf{59.2}\pm4.8$ & $\textbf{7.2}\pm3.1$  \\  
   
\bottomrule
\end{tabular}
 
\end{table}

\subsection{Thermal Sampling vs.\ Encoding}\label{sec:svse}
By design, $T^{3}S$ couples thermal-time sampling with cloud-aware selection,
retaining the least-cloudy acquisition in each thermal-time bin, so cloud filtering
is intrinsic to the method. To verify that its gains stem from thermal-time
alignment rather than cloud filtering alone, we build a strengthened baseline,
DS+TPE: we equip the encoding-based approach of~\cite{nyborg2022generalized} (TPE)
with the same cloud-aware in-bin selection (stronger than its original random
sampling), so the two differ only in whether the thermal signal is injected
through \emph{sampling} ($T^{3}S$) or \emph{encoding}.
Even against this stronger baseline, $T^{3}S$ outperforms DS+TPE in accuracy
($77.0$ vs.\ $76.4$) and calibration ($1.1\%$ vs.\ $6.3\%$ ECE, approximately
$6\times$ lower). It also shows lower fold-to-fold variation, indicating more
robust cross-year behavior. The advantage further widens in operationally
challenging settings: under early-season prediction, $T^{3}S$ improves accuracy
and calibration by $+4.0$ accuracy points and $-8.7$ ECE points, while under
low-label training it gives $+3.3$ accuracy points and $-2.8$ ECE points
(Table~\ref{tab:ds_tpe}). 
Finally, because $T^{3}S$ operates at the input level, it is not tied to a modifiable positional-encoding module. It therefore applies without architectural changes in settings where TPE cannot: (i) CNN- and RNN-based backbones that lack a positional encoding to modify; (ii) embedding-based pipelines such as Tessera~\cite{feng2026tessera}, where the encoding is fixed; and (iii) models with coarse temporal granularity, where injecting fine-grained thermal information via TPE would require costly re-training.
For further details and discussion, see Appendix~\ref{app:ts_vs_tpe}.

\section{Conclusion}\label{conclusion}

We present Thermal Time-based Temporal Sampling ($T^{3}S$), a model-agnostic approach that leverages cumulative growing degree days to subsample satellite time series in a phenologically meaningful way. Evaluated on the SwissCrop dataset across six cross-year folds and two cross-region transfer directions using three different backbones, $T^{3}S$ consistently outperforms state-of-the-art baselines in both classification accuracy and uncertainty calibration, an aspect often overlooked in the literature.
Alongside $T^{3}S$, we publicly release SwissCrop, the first country-scale, multi-year crop dataset with paired temperature data, to facilitate reproducible research on phenology-aware and cross-year crop mapping.
The strength of $T^{3}S$ lies in its induced inductive bias: by re-indexing observations by thermal time rather than calendar days, phenologically equivalent growth stages are consistently aligned across years, reducing the input distribution shift that undermines calendar-based models. This steers the model toward meaningful phenological signals, shaping more informative representations and producing more accurate predictions and well-calibrated uncertainty estimates.

\paragraph{\textbf{Limitation and Future Work.}}
Despite consistent gains across multiple datasets and backbones, $T^3S$ relies on
temperature as the sole phenological driver, whereas other factors such as precipitation,
photoperiod, and irrigation also influence crop development, particularly in tropical
regions; incorporating such variables is a natural next step. We have so far evaluated
$T^3S$ on a limited set of backbones (U-TAE, PSE+LTAE, Galileo), and extending it to
further architectures, especially large pretrained Earth Observation foundation models,
is a promising direction. An interesting case is applying $T^3S$ directly on top of
embedding-based representations, e.g., Tessera~\cite{feng2026tessera}, where thermal-time re-indexing is directly applicable. 
Finally, while $T^3S$ and TPE appear largely redundant on U-TAE (Appendix~\ref{app:pe_ablation}), they operate at different levels (sampling vs. encoding) and could be complementary in attention architectures that rely more heavily on positional information.

\bibliographystyle{splncs04}
\bibliography{main}

\clearpage
\section*{\Large Appendix}

\appendix

\renewcommand{\theHsection}{appendix.\Alph{section}}
\renewcommand{\theHsubsection}{appendix.\Alph{section}.\arabic{subsection}}
\renewcommand{\theHsubsubsection}{appendix.\Alph{section}.\arabic{subsection}.\arabic{subsubsection}}

\section{Preliminary Analysis with U-TAE on PASTIS}
\label{app:prelim}

Before designing $T^3S$, we analyze two temporal design choices in U-TAE to motivate our
sampling strategy. To avoid biasing the design toward our evaluation data, we conduct this
analysis on PASTIS~\cite{garnot2021panoptic}, a well-established single-year crop benchmark
introduced alongside U-TAE. PASTIS mirrors SwissCrop's multi-temporal structure while
remaining fully independent, enabling controlled analysis without leaking information from the
target dataset. Each PASTIS series contains, on average, 48 Sentinel-2 observations (up to 61).

\paragraph{\textbf{Temporal length.}}
We first uniformly subsample each sequence to $T=24$ time steps. As shown in
Table~\ref{tab:pastis_temp}, this yields only a $0.2\%$ absolute drop in overall accuracy
while roughly halving GPU memory and multiply–accumulate operations (MACs) per forward pass.
We therefore fix $T=24$ in all subsequent experiments.

\begin{table}[h]
\centering
\caption{Performance comparison for varying temporal length on the PASTIS dataset. Results are averaged over a 5-fold experimental setting. Refer to \cite{garnot2021panoptic} for details.}
\label{tab:pastis_temp}
\begin{tabular}{c | c | c | c | c | c }
\toprule
\textbf{$T_{min}$} & \textbf{$T_{max}$} & \textbf{$T_{mean}$} &  \textbf{Memory (Gb)} &  \textbf{MACs (B)} &  \textbf{Accuracy (\%)} \\
\midrule
  38 & 61 & 48 & 16.8 & 288 & $83.1\pm 0.5$   \\
\midrule
  24 & 24 & 24 & 8.0 & 145 & $82.9\pm 0.5$  \\              
\bottomrule
\end{tabular}
 
\end{table}

\begin{figure*}[t]
    \centering
    \includegraphics[width=0.245\textwidth]{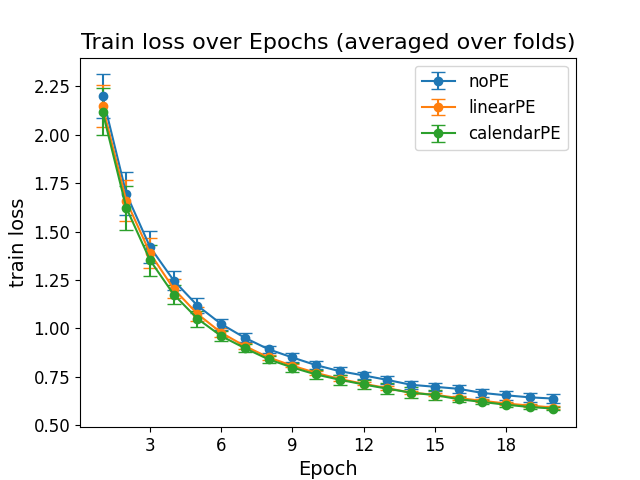}%
    \hfill
    \includegraphics[width=0.245\textwidth]{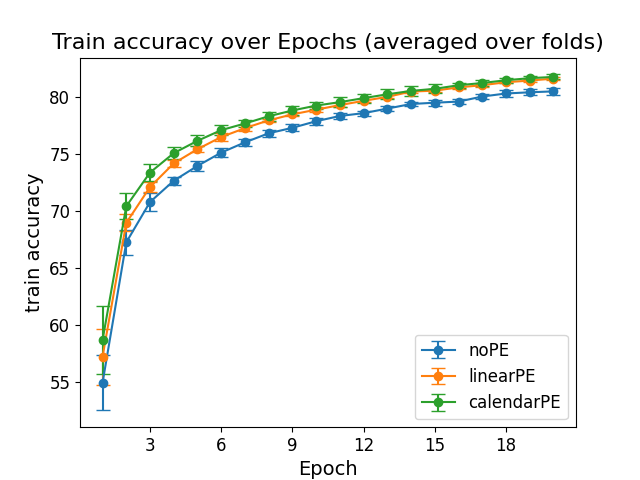}%
    \hfill
    \includegraphics[width=0.245\textwidth]{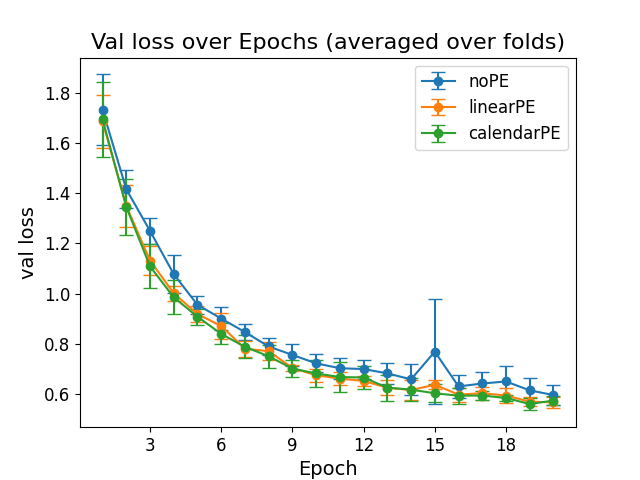}%
    \hfill
    \includegraphics[width=0.245\textwidth]{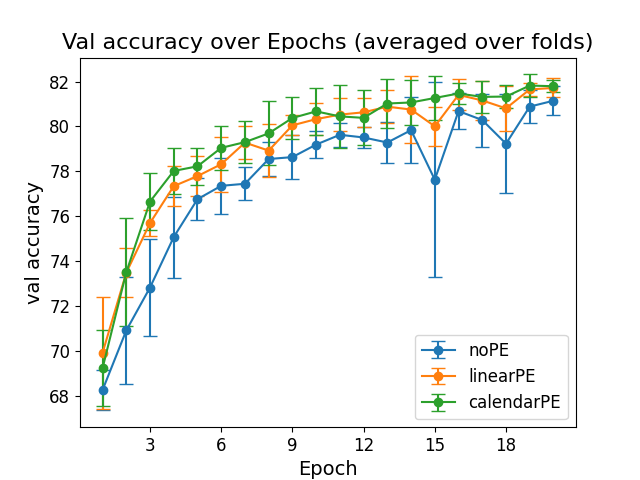}%
    \caption{Learning curves of the U-TAE model with different position encodings (PE) on the
    PASTIS dataset, showing training loss, training accuracy, validation loss, and validation
    accuracy, respectively. linearPE and calendarPE converge to nearly identical results, while
    removing PE degrades performance.}
    \label{fig:pastis}
\end{figure*}

\paragraph{\textbf{Positional encoding.}}
We then add a standard sinusoidal positional encoding (PE) to the input embeddings to mark
temporal order, and experiment with three variants: \emph{noPE}, where all position embeddings
are set to zero; \emph{linearPE}, where positions are encoded as sequential indices
$(1,2,3,\dots)$; and \emph{calendarPE}, where positions are encoded as the actual calendar
day-of-year (e.g., $2,5,\dots,365$). Removing PE degrades performance, yet linearPE and
calendarPE converge to nearly identical results (Figure~\ref{fig:pastis}), indicating that
U-TAE relies mainly on relative ordering rather than absolute time values. Since uniform
subsampling already imposes this ordering, altering the encoding on top adds little; this does
not rule out PE in other settings, but it shows that, given an ordered subsample, the encoding
is largely redundant with U-TAE backbone.

In summary, (i) the temporal sequence can be compressed with little loss in accuracy while
yielding a more efficient model, and (ii) temporal order matters while absolute timestamps add
limited benefit. These observations motivate selecting a compact set of phenologically
informative observations rather than modifying the positional encoding.

\section{Class-wise Performance Comparison}
\begin{figure}[h]
    \centering
    \includegraphics[width=1.0\textwidth]{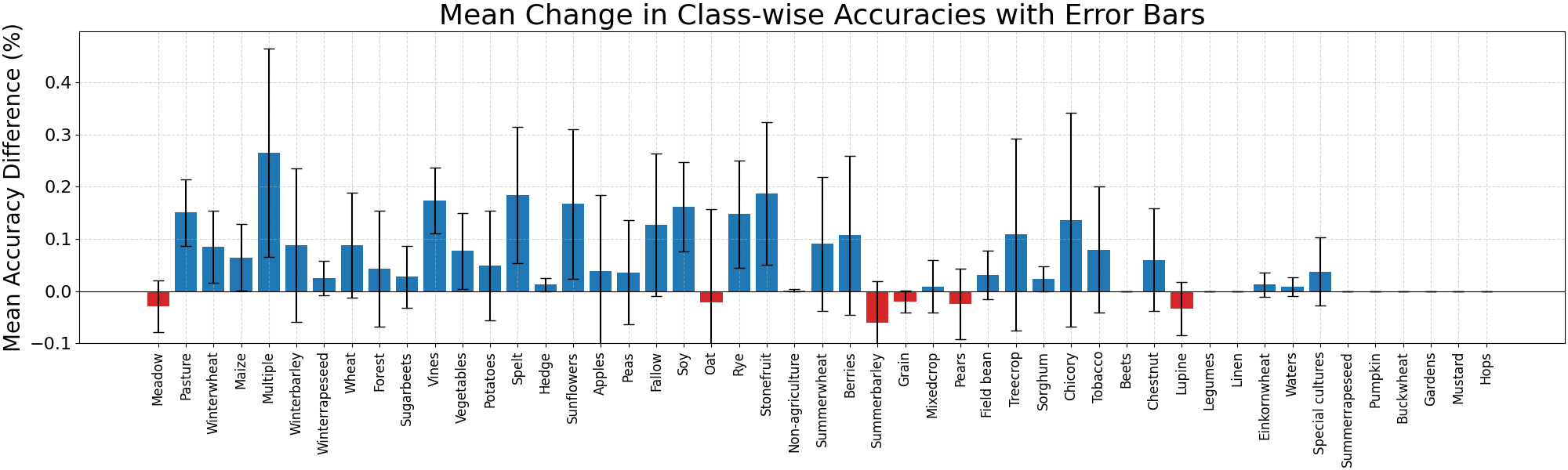}
    \caption{Class-wise accuracy difference between $T^{3}S$ and U-TAE. Averaged over six experimental folds. Blue bars indicate classes where $T^{3}S$ outperforms U-TAE, while red bars indicate classes where $T^{3}S$ underperforms.}
    \label{fig:classwise}
\end{figure}

Figure~\ref{fig:classwise} shows the per-class accuracy difference between $T^{3}S$ and U-TAE, averaged over six folds. $T^{3}S$ improves accuracy for the vast majority of crop types, with the largest gains observed for mid-frequency classes. Only the rarest classes (e.g., hops) remain at zero due to insufficient training samples, and meadow accuracy drops slightly, reflecting U-TAE's tendency to overpredict dominant classes.

\section{Sensitivity Analysis on Temporal Length}\label{app:temp}
\begin{figure}[h]
    \centering
    \includegraphics[width=0.5\textwidth]{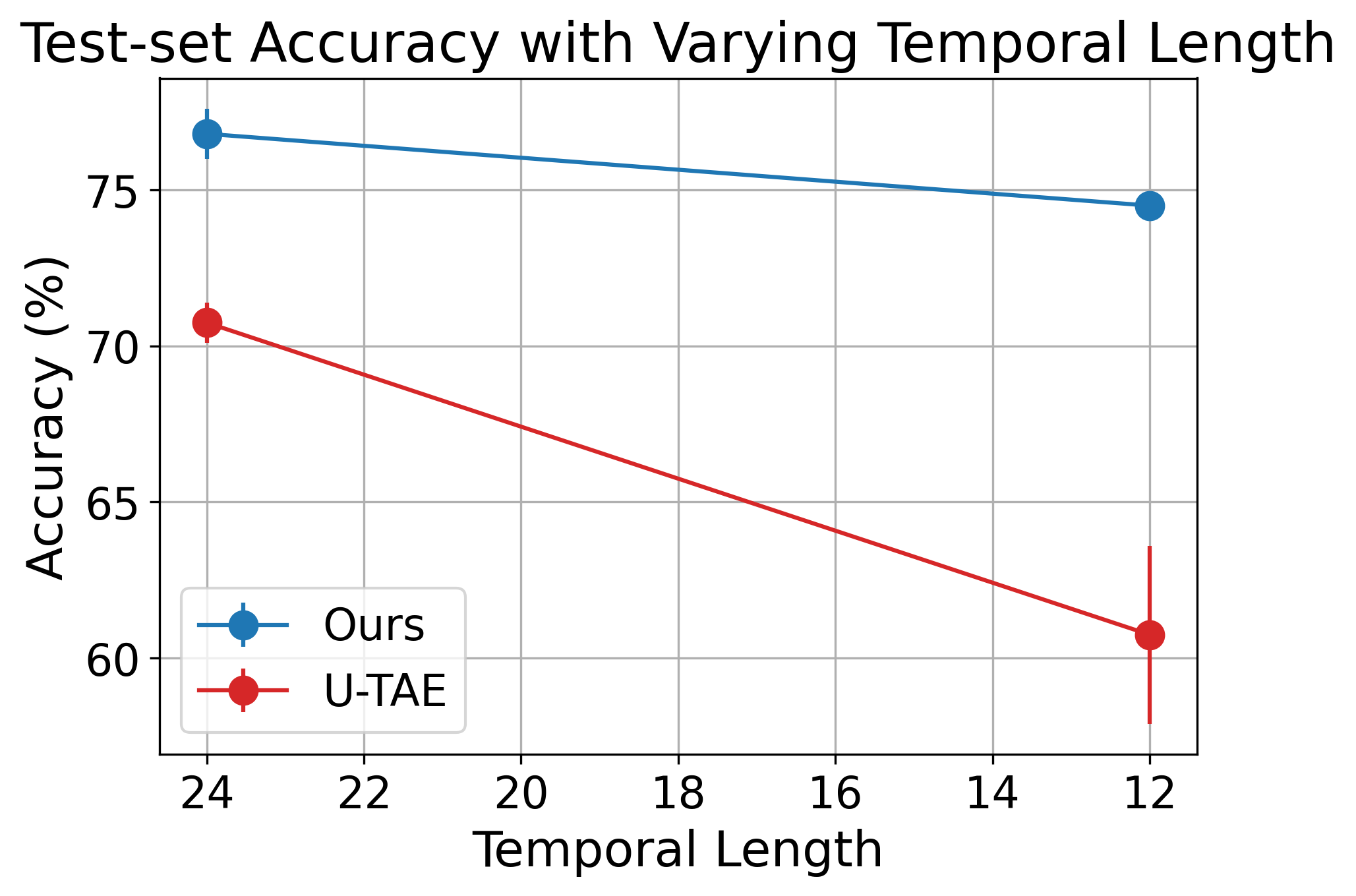}%
    \includegraphics[width=0.5\textwidth]{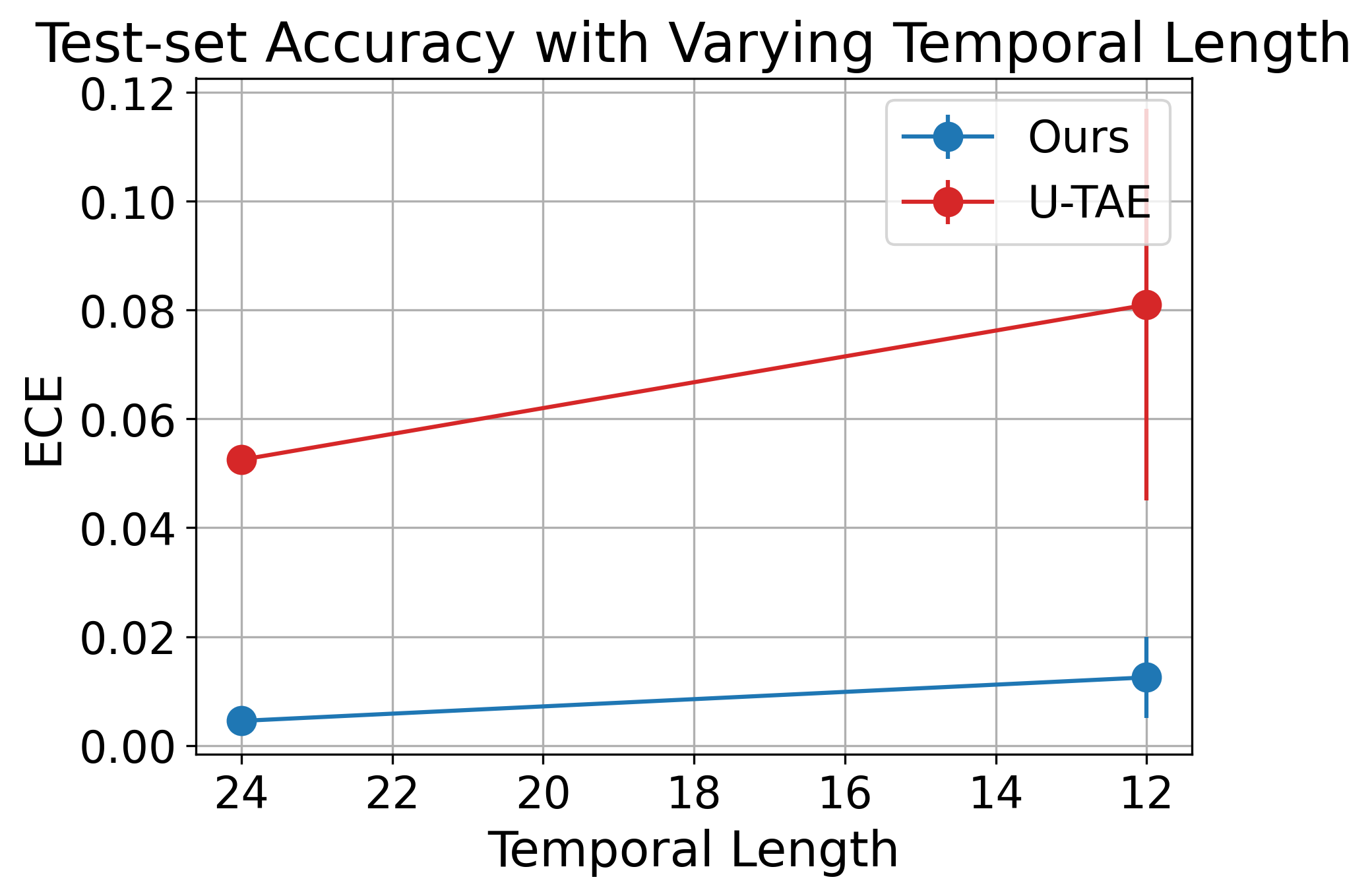}%
\caption{Sensitivity to temporal sequence length. Models are trained on the 2021 dataset and evaluated on the 2022 and 2023 datasets. Results are averaged over both test folds.}
    \label{fig:tempLen}
\end{figure}

We evaluate the effect of temporal sequence length by comparing models trained with 24 and 12 sampled timestamps. As shown in Figure~\ref{fig:tempLen}, reducing the sequence length from 24 to 12 leads to only a minor drop in accuracy and calibration for $T^{3}S$, while U-TAE shows a larger decline. Notably, $T^{3}S$ with just 12 timestamps still significantly outperforms U-TAE with the full 24 timestamps, demonstrating that thermal-time sampling efficiently captures key phenological phases even with half the sequence length. In this experiment, the models are trained on 2021 and tested on the remaining years.

\section{Ablation on Positional Encoding}
\label{app:pe_ablation}

To verify that $T^{3}S$ does not depend on a specific positional encoding scheme, we compare $T^{3}S$ with and without thermal-time positional encoding on the SwissCrop dataset (trained on 2021, evaluated on 2022 and 2023). As shown in Table~\ref{tab:pe_ablation}, removing positional encoding results in a negligible drop in accuracy while noticeably improving NLL, suggesting that thermal-time sampling already provides sufficient temporal structure and that adding thermal PE introduces marginal redundancy. Based on this finding, we omit positional encoding in all experiments.


\begin{table}[ht]
\centering
\caption{Ablation on positional encoding for $T^{3}S$. Models are trained on 2021 and evaluated on 2022 and 2023. Results are averaged over both test folds. Best score for each metric in \textbf{bold}.}
\label{tab:pe_ablation}
\begin{tabular}{lcc}
\toprule
\textbf{Method} & \textbf{Acc(\%) ($\uparrow$)} & \textbf{NLL ($\downarrow$)} \\
\midrule
$T^{3}S$ + Thermal PE & \textbf{76.9} & 0.764 \\
$T^{3}S$ (no PE)      & 76.8 & \textbf{0.706} \\
\bottomrule
\end{tabular}
\end{table}

\section{Thermal-Time Sampling vs.\ Thermal-Time Positional Encoding}
\label{app:ts_vs_tpe}

$T^3S$ and thermal-time positional encoding (TPE)~\cite{nyborg2022generalized} share the
same goal -- injecting phenological bias via cumulative growing degree days -- but differ
in \emph{how} and \emph{where} they act. We first isolate the effect of the injection
mechanism through a controlled ablation, then discuss the architecture-level distinctions
that follow from operating at the input level rather than the encoding level.

\subsection{Sampling vs.\ Encoding: A Controlled Ablation}
\label{app:ds_tpe_ablation}

As noted in the main text, $T^3S$ couples thermal-time sampling with cloud-aware selection:
within each thermal-time bin it retains the least-cloudy acquisition, so cloud filtering is
intrinsic to the method. Since cloud filtering independently improves accuracy, a fair
assessment of the thermal-time \emph{sampling} contribution must hold it fixed. We therefore
construct a strengthened baseline, Deformable Sampling with thermal-time positional encoding
(DS+TPE), by equipping the encoding-based approach of~\cite{nyborg2022generalized} with the
\emph{same} cloud-aware in-bin selection used by $T^3S$. The two then differ only in whether
the thermal signal is injected through \emph{sampling} (thermal-time bins, $T^3S$) or through
\emph{encoding} (calendar-time bins + TPE), which isolates the injection mechanism. Note that
this baseline is stronger than the original TPE of~\cite{nyborg2022generalized}, which samples
randomly within each interval and thus confounds thermal-time encoding with cloud filtering.

Table~\ref{tab:ds_tpe} reports results on SwissCrop, averaged over the six cross-year folds.
Even against this strengthened baseline, $T^3S$ outperforms DS+TPE on both accuracy and
calibration, with roughly $6\times$ lower ECE ($1.1\%$ vs.\ $6.3\%$) and substantially better
fold-to-fold stability (Acc std $0.5$ vs.\ $1.2$; ECE std $0.6$ vs.\ $2.0$). Notably, DS+TPE's
calibration ($6.3\%$ ECE) is \emph{worse} than vanilla U-TAE ($5.1\%$, Table~\ref{tab:performance}):
adding TPE raises accuracy at the cost of calibration, whereas $T^3S$ improves both. DS+TPE
remains competitive when train and test years are climatologically similar
(e.g., 2022$\rightarrow$2023) but degrades under stronger shifts (e.g., 2021$\rightarrow$2022),
indicating that thermal-time encoding alone cannot fully realign the input distribution.

The gap widens under challenging operational conditions. On the 2021$\rightarrow$2022 fold with
an end-of-June cutoff, $T^3S$ reaches $68.2\%$ / $3.4\%$ (Acc / ECE) versus DS+TPE's $64.2\%$ /
$12.1\%$ ($+4.0\%$ Acc, $-8.7\%$ ECE); with only $10\%$ of training labels, $T^3S$ achieves
$72.3\%$ / $1.5\%$ versus $69.0\%$ / $4.3\%$ ($+3.3\%$ Acc, $-2.8\%$ ECE). Together, these
results confirm that thermal-time \emph{sampling}, not encoding, is the more effective mechanism
for injecting phenological bias, particularly where reliable calibration matters most.

\begin{table}[t]
\centering
\caption{Thermal-time sampling ($T^3S$) vs.\ thermal-time positional encoding (DS+TPE) on
SwissCrop. Both use cloud-aware in-bin selection, isolating the injection mechanism. Average
over six cross-year folds; hard cases on the 2021$\rightarrow$2022 fold. Best per metric in
\textbf{bold}.}
\label{tab:ds_tpe}
\begin{tabular}{llcc}
\toprule
Setting & Method & Acc (\%) $\uparrow$ & ECE (\%) $\downarrow$ \\
\midrule
\multirow{2}{*}{Average (6 folds)}
 & DS + TPE      & $76.4 \pm 1.2$ & $6.3 \pm 2.0$ \\
 & $T^3S$ (ours) & $\mathbf{77.0 \pm 0.5}$ & $\mathbf{1.1 \pm 0.6}$ \\
\midrule
\multirow{2}{*}{Early-season (end of June)}
 & DS + TPE      & $64.2$ & $12.1$ \\
 & $T^3S$ (ours) & $\mathbf{68.2}$ & $\mathbf{3.4}$ \\
\midrule
\multirow{2}{*}{Low-data ($10\%$ labels)}
 & DS + TPE      & $69.0$ & $4.3$ \\
 & $T^3S$ (ours) & $\mathbf{72.3}$ & $\mathbf{1.5}$ \\
\bottomrule
\end{tabular}
\end{table}

\subsection{Architecture-Level Distinctions}
\label{app:tpe_arch}

Beyond performance, $T^3S$ and TPE differ in applicability because they act at different
levels. TPE replaces the model's positional-encoding module (architecture level), whereas
$T^3S$ re-indexes observations before encoding (input level / preprocessing). Operating at
the input level, $T^3S$ applies in three settings where TPE cannot:
\begin{enumerate}
\item \textbf{Architectures without a positional-encoding module}, such as RNN- or
CNN-based backbones, where there is no PE to modify.
\item \textbf{Models with coarse temporal granularity}, e.g.\ monthly timesteps, where
injecting fine-grained thermal information via TPE would require retraining.
\item \textbf{Frozen pretrained foundation models} (e.g., Tessera, AlphaEarth) used with a
lightweight downstream classifier, where the backbone's PE cannot be altered.
\end{enumerate}
Because it requires no modifiable PE, $T^3S$ is applicable in all these settings and is
complementary to TPE where a modifiable PE is available.

\section{SwissCrop Dataset Details}\label{app:data}



\begin{figure}[t]
    \centering
    \includegraphics[width=1.0\textwidth]{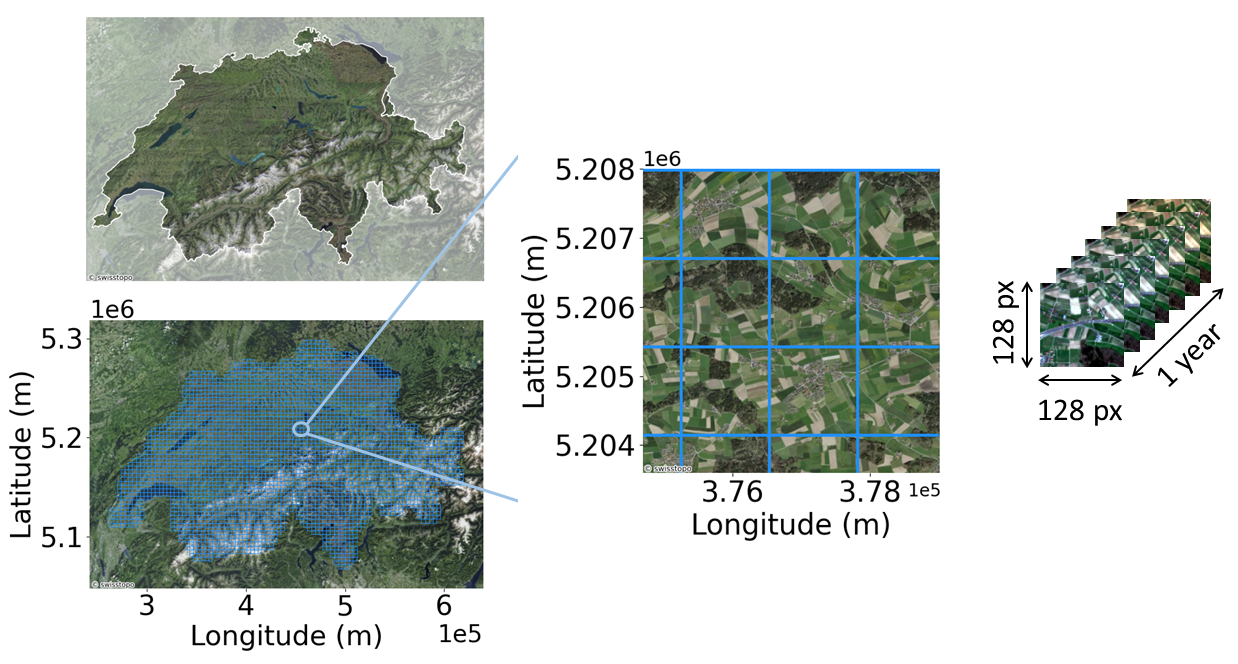}
    \caption{Region of interest (top left), the SwissCrop dataset (bottom left) and its coverage by data cubes (center), with an example of a single analysis-ready cube (right).}
    \label{fig/dataset_overview}
\end{figure}

\begin{figure*}[th]
    \centering
    \includegraphics[width=1.0\textwidth]{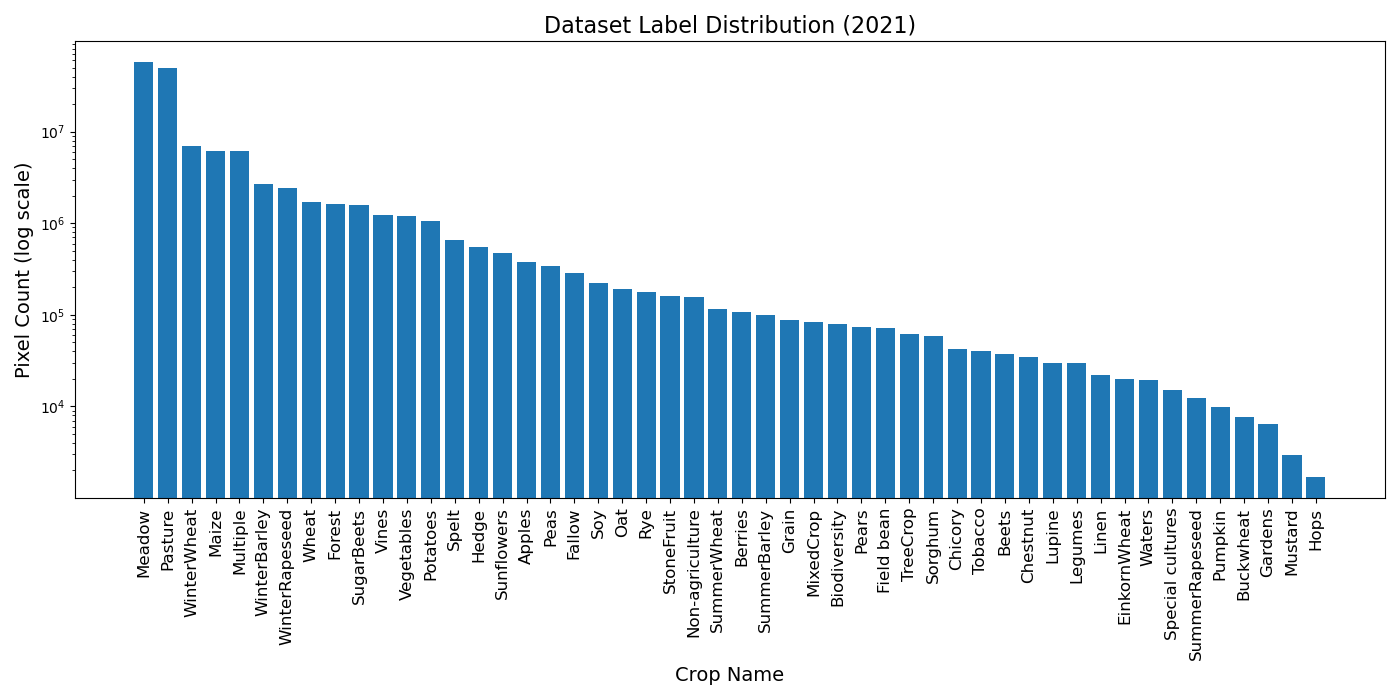}
    \caption{SwissCrop dataset label distribution for the year 2021. Note the logarithmic scale of the y-axis.}
    \label{fig:label_dist}
\end{figure*}

In this work, we work with \textbf{SwissCrop 2021, 2022, 2023}, a three-year crop dataset covering the entirety of Switzerland (see Figure~\ref{fig/dataset_overview}) from 2021 to 2023. 
SwissCrop combines Sentinel-2 Level-2A bottom-of-atmosphere multi-spectral imagery, averaging 110 timestamps per season, with annual crop‐type labels for each field.
The dataset includes 50 distinct crop types, each representing the primary crop grown in a field during a given season. 
These labels were provided by farmers to the cantons of Switzerland according to a data model provided by the Federal Office for Agriculture.\footnote{https://www.blw.admin.ch/de/landwirtschaftliche-kulturflaechen}
%
%
Unlike many existing datasets, e.g. PASTIS dataset \cite{garnot2021panoptic}, SwissCrop faithfully reflects the pronounced class imbalance found in real‐world farming: a long‐tail distribution in which a handful of major crops dominate while the majority occur only sporadically (see Figure~\ref{fig:label_dist}).

The data is organized into analysis-ready data cubes optimized for deep learning applications (Figure~\ref{fig/dataset_overview}). Each cube corresponds to a single year and covers a 128×128 pixel region (equivalent to 1280×\qty{1280}{\meter} at Sentinel-2’s \qty{10}{\meter} resolution). We include the following spectral bands in each cube: B02 (Blue), B03 (Green), B04 (Red), B05 (Vegetation Red Edge1), B06 (Vegetation Red Edge2), B07 (Vegetation Red Edge3), B08 (Near-Infrared), B8A (Red Edge4), and B12 (Short-Wave Infrared2). 
All imagery is reprojected to UTM zone 32N and resampled so that pixel coordinates align to a \qty{10}{\meter} grid. In addition to optical bands, each cube stores the Scene Classification Layer (SCL), which provides both land-cover classification and cloud information. To minimize storage requirements, band values are saved as 16-bit integers, representing reflectance scaled by 10,000. Over the three years, SwissCrop comprises around 20K cubes per year, fully covering Switzerland’s agricultural area with consistent preprocessing and labeling. This dataset enables large-scale, cross-year experiments in crop classification and phenology analysis.

To complement these remote sensing observations, we incorporate high-resolution gridded climate data from MeteoSwiss, which supplies near-surface temperature fields (minimum, maximum, and average) on 1 km, 2 km, and 5 km grids \cite{federal_office_of_meteorology_and_climatology_meteoswiss_climate_2024}. In our study, we use the 1 km daily minimum (TminD) and maximum (TmaxD) temperature series to compute growing degree days (GDD; refer to Section~\ref{method:t3s}).


We make the full SwissCrop dataset publicly available to the research community. We hope this resource will facilitate further advances in large-scale, cross-year crop classification, phenology modeling, and earth observation research.

\begin{table}[th]
\centering
\caption{Crop type mapping benchmark datasets.
\emph{Type}: parcel-aggregated time series (TS) vs. satellite image time series (SITS).
\emph{Classes}: number of agricultural classes.
\emph{Multi-year}: \checkmark for repeated observations of the same units across years, (\checkmark) for multi-year coverage over different units.
\emph{Temp}: temperature data paired with imagery.}
\label{tab:datasets}
\setlength{\tabcolsep}{4pt}
\resizebox{1\linewidth}{!}{
\begin{tabular}{lllllcc}
\toprule
\textbf{Dataset} & \textbf{Type} & \textbf{Scope} & \textbf{Classes} & \textbf{\# Years} & \textbf{Multi-year} & \textbf{Temp}\\
\midrule
BreizhCrops~\cite{breizhcrops2020}           & TS   & Regional (FR)  & 9      & 1\phantom{*} \hspace{0.3em}(2017)          & ---          & ---          \\
TimeSen2Crop~\cite{weikmann2021timesen2crop} & TS   & National (AT)  & 16     & 2\phantom{*} \hspace{0.3em}(2018--2019)     & \checkmark   & ---          \\
EuroCropsML~\cite{eurocropsml2025}           & TS   & Transnational  & 176    & 1\phantom{*} \hspace{0.3em}(2021)           & ---          & ---          \\
TimeMatch~\cite{nyborg2022timematch}         & TS   & Transnational  & 16     & 1\phantom{*} \hspace{0.3em}(2017)           & ---          & \checkmark   \\
CropDeepTrans~\cite{BARRIERE2024114110}      & TS   & Transnational  & 151    & 5\phantom{*} \hspace{0.3em}(2016--2020)     & \checkmark   & ---          \\
\midrule
MunichCrops~\cite{Russwurm2018}              & SITS & Regional (DE)  & 17     & 2\phantom{*} \hspace{0.3em}(2016--2017)     & (\checkmark) & ---          \\
PASTIS~\cite{garnot2021pastis}               & SITS & Regional (FR)  & 18     & 1\phantom{*} \hspace{0.3em}(2019)           & ---          & ---          \\
ZueriCrop~\cite{turkoglu2021crop}            & SITS & Regional (CH)  & 48     & 1\phantom{*} \hspace{0.3em}(2019)           & ---          & ---          \\
DENETHOR~\cite{kondmann2021denethor}         & SITS & Regional (DE)  & 9      & 2\phantom{*} \hspace{0.3em}(2018--2019)     & \checkmark   & ---          \\
Sen4AgriNet~\cite{sykas2022sen4agrinet}      & SITS & Transnational  & 158    & 2* \hspace{0.3em}(2016--2020)               & (\checkmark) & ---          \\

\midrule
\textbf{SwissCrop} & \textbf{SITS} & \textbf{National (CH)} & \textbf{50} & \textbf{3\phantom{*} (2021--2023)} & \textbf{\checkmark} & \textbf{\checkmark} \\
\bottomrule
\multicolumn{7}{r}{\footnotesize $^*$only 2019--2020 are publicly available.} \\
\end{tabular}
}
\end{table}

\subsection*{Comparison with Existing Crop Mapping Datasets}
\label{app:datasets}

Table~\ref{tab:datasets} positions SwissCrop against widely used crop mapping benchmarks along
five axes: input type, spatial scope, number of agricultural classes, temporal coverage, and
paired temperature data. We distinguish parcel-aggregated time series (TS), which store one
sequence per field, from satellite image time series (SITS), which preserve full spatial
structure and support dense segmentation.

Two properties are central to cross-year phenology-aware evaluation. The first is multi-year
coverage of the same spatial units, which enables leave-one-year-out testing under genuine
inter-annual variability; datasets that span multiple years over different regions
(\checkmark) support this only partially, since spatial and temporal shifts are confounded.
The second is paired temperature data, required to compute growing degree days and re-index
observations onto a thermal axis. Among existing datasets, only TimeMatch provides temperature,
and it covers a single year, so it cannot be used for cross-year thermal-time evaluation.

SwissCrop is the only dataset that combines national spatial scope, repeated multi-year
coverage of fixed spatial units, dense SITS structure, and paired temperature data. This
combination makes it uniquely suited to benchmark phenology-aware methods such as T$^3$S under
realistic cross-year conditions.

\section{Training Details}\label{training_details}

The model was trained for 100 epochs using a batch size of 8 and a fixed temporal length ($T$) of 24.
The input image size ($H$ and $W$) is set to $128 \times 128$ pixels.
 We employed the Adam optimizer with an initial learning rate of 0.001, scheduled by a OneCycleLR policy with a maximum learning rate of 0.01 and a warm-up phase covering 30\% of the total training steps. 
%
%
The U-TAE architecture is configured with an encoder having channel widths \([128, 128, 128, 256]\), a symmetric decoder with widths \([32, 64, 128, 256]\) and strided convolutions with kernel size \(k=4\), stride \(s=2\), and padding \(p=1\). The temporal attention module employs \(n_{\text{head}}=16\) heads and a model dimension of \(d_{\text{model}}=256\). Input features are standardized to zero mean and unit variance on a per-channel basis. We implemented the model in PyTorch and trained it on a single NVIDIA RTX 4090 GPU over roughly three days. For reproducibility of experiments, the full codebase and pretrained model checkpoints will be publicly available on GitHub.




\end{document}